%% file: main.tex
\documentclass{article}
\usepackage{iclr2027_conference,times}
\usepackage[T1]{fontenc}
\usepackage{amsmath,amssymb,booktabs,graphicx,longtable,array,multicol,ragged2e,float,placeins}
\usepackage[hidelinks]{hyperref}
\usepackage{tikz}
\usetikzlibrary{arrows.meta,positioning}
\graphicspath{{figures/}}
\title{FORESIGHT-9: Prospective and Process-Aware Evaluation of Adaptive Trading Agents}
\author{Xiangxin Luo\textsuperscript{1}, Chengtian Hong\textsuperscript{1}, Haohua Li\textsuperscript{2}, Yongyi Xie\textsuperscript{3}\\
\textsuperscript{1}University of Science and Technology of China\\
\textsuperscript{2}University of Hong Kong \quad
\textsuperscript{3}Shanghai Jiao Tong University}

\providecommand{\Description}[1]{}
\iclrfinalcopy
\begin{document}
\maketitle
\begin{abstract}
Trading-agent evaluation can fail at three successive levels: a ranking on
realized history may not transfer reliably to alternative future paths, prospective returns may
not exceed simple non-learning controls, and a favorable terminal outcome may
persist after adaptation itself has stopped. We introduce FORESIGHT-9, a
prospective and process-aware benchmark with nine auditable counterfactual
worldlines branching from a common July 2026 information boundary. A common
contract fixes the market panel, cadence, constraints, costs, and account writes
while preserving framework-specific research and proposal mechanisms. Across
36 long-horizon runs from two frameworks and two model backbones,
Native-Policy Retrospective Replay yields a historical ranking whose median
Spearman correlation with prospective rankings is $-0.8$; its winner remains
first in only $1/9$ worldlines. Prospectively, periodic equal weighting and a
causal inverse-volatility rule each outperform 31 of 36 agent--worldline runs,
showing that positive agent returns need not establish incremental value from
online adaptation. Finally, in a high-return WL8 run the live research library
collapses while the declared ensemble persists and executed holdings converge
toward equal weight: the run keeps earning after meaningful self-evolution has
ceased. FORESIGHT-9 therefore evaluates future robustness, incremental value
over simple controls, and whether the intended adaptive mechanism remains
functional. We release the worldlines, trajectories, audit traces, and
deterministic regeneration scripts.
\end{abstract}
\input{sections/introduction}
\input{sections/related_work}
\input{sections/worldlines}
\input{sections/agents}
\input{sections/results}
\input{sections/limitations}
\input{sections/conclusion}
\clearpage
\section*{AI Use Statement}
Generative AI tools were used to assist with literature retrieval and discovery,
including identifying related work and existing benchmark references, and to
generate and cross-review the synthetic macro-financial scenario
specifications underlying FORESIGHT-9. Generative AI tools were also used to
assist with code for scientific visualization and with the creation or
modification of scientific figures. The research methodology, experimental
execution, interpretation of results, and manuscript text were carried out by
the authors. All AI-assisted datasets, code, figures, citations, and retrieved
sources were reviewed and validated by the authors, who take full
responsibility for the paper and accompanying artifacts.

\section*{Reproducibility Statement}
The supplement releases the nine market panels, processed trajectories,
learned-policy boundary artifacts, common execution code, frozen parameters,
per-run prospective metrics, and package-relative scripts for worldline and
factor-inventory regeneration, RR reconstruction, deterministic verification,
summary regeneration, and quantitative figures.
Appendices~\ref{app:worldlines}, \ref{app:rr}, and \ref{app:provenance} specify
the data, replay contract, and provenance boundaries.
\section*{Ethics Statement}
All experiments are conducted offline. Prospective evaluation uses synthetic
counterfactual worldlines, while Native-Policy Retrospective Replay uses
realized historical data; no live-market execution or client funds are
involved. FORESIGHT-9 is a research instrument, not investment advice.
\clearpage
\bibliographystyle{plainnat}
\bibliography{references}
\appendix
\input{sections/appendix_nav}
\input{sections/appendix_metrics}
\input{sections/appendix_worldlines}
\input{sections/appendix_significance}
\input{sections/appendix_rr}
\input{sections/appendix_factors}
\input{sections/appendix_process}
\input{sections/appendix_provenance}
\input{sections/appendix_models}
\end{document}

%% file: sections/introduction.tex
\section{Introduction}
\label{sec:intro}

Foundation-model agents are increasingly used for financial research,
factor discovery, and portfolio construction, yet they are still
evaluated primarily through retrospective backtests on realized market
history. This habit leaves a three-level identification problem. First, a
historical leaderboard cannot show whether the same ordering survives across
future worldlines \citep{glasserman2023lookahead,didisheim2025predictable,
zhu2026ktdfin,white2000reality,bailey2017pbo}. Second, even a favorable
prospective return does not show that a complex adaptive agent added value
beyond passive exposure, mechanical rebalancing, or a simple causal risk
rule. Third, an agent's terminal return does not show that its intended
adaptation remained functional: the live research state, declared decision
state, and executed portfolio can diverge while the account continues to earn.

\begin{figure}[t]
\centering
\includegraphics[width=\columnwidth]{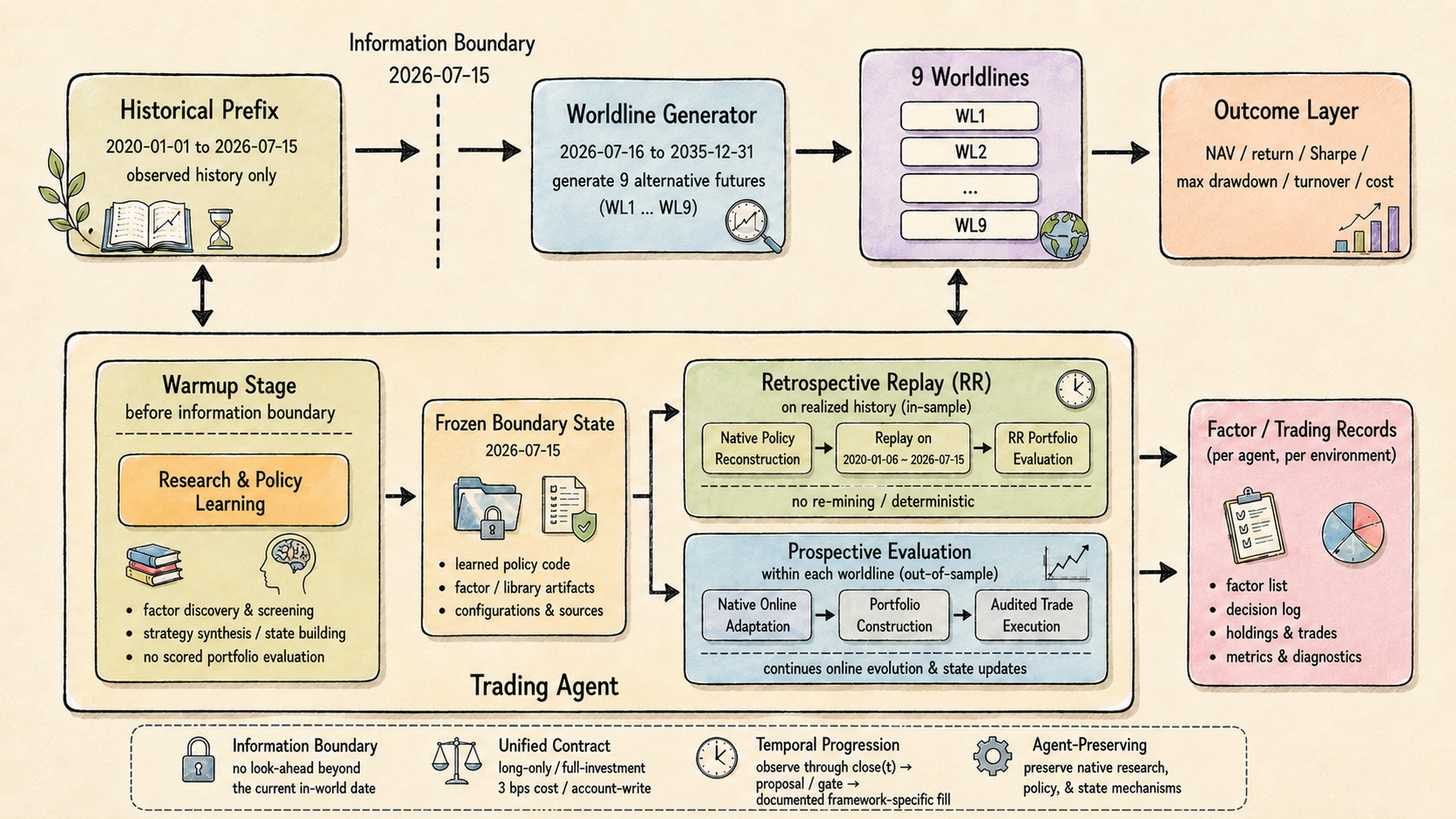}
\caption{\textbf{FORESIGHT-9 overview.} The shared pre-boundary warmup
produces the learned research and policy state without scored portfolio
evaluation. At the information boundary, the frozen state supports both the
deliberately in-sample Retrospective Replay (RR) diagnostic and prospective
online adaptation across nine worldlines. Outcomes and factor/trading records
are audited separately, while the contract fixes temporal progression,
portfolio constraints, costs, and the account-write interface. In the figure, 2026-07-16 is the shared
observed execution/market anchor and generator handoff; scenario-specific
synthetic rows begin on 2026-07-17.}
\label{fig:overview}
\end{figure}

We present \textbf{FORESIGHT-9}, a prospective benchmark that evaluates
foundation-model trading agents across nine auditable counterfactual
futures with staged macro-financial events, multi-asset anchors, and
time-gated observations (\S\ref{sec:worldlines}). Agents observe only
what each in-world date discloses and must search for, validate, and
admit factors, rank signals, construct portfolios, and execute under a
controlled observation--proposal--execution interface with audited
framework-specific fills while evolving their factor libraries and research memories (\S\ref{sec:agents}). The contract standardizes the market panel, cadence, constraints, costs,
and failure handling, but deliberately preserves each agent's native search
and self-evolution mechanisms---a benchmark should not erase the behavior it
intends to measure. Figure~\ref{fig:overview} summarizes the evaluation
pipeline.

To directly test whether the conclusion induced by realized-history
evaluation transfers to unseen futures, we freeze each configuration's
learned policy at the common boundary and replay its framework-specific
native policy over the same 2020--2026 history. This Native-Policy
Retrospective Replay (RR) executes the learned signal and target-weight
proposal mechanism from archived native artifacts, while evaluating all
proposals under a common causal close$(t)\!\to$open$(t+1)$ execution contract.
``Native'' refers to the learned policy and proposal-generation mechanism, not
to end-to-end execution semantics. RR is deliberately in-sample and is used
only as a ranking diagnostic, not as an estimate of future performance.

FORESIGHT-9 makes three contributions.
\begin{itemize}
\item \textbf{Prospective multi-worldline evaluation.} We replace the
  single realized test path with nine controlled counterfactual stress
  futures branching from a shared information boundary. Each worldline
  combines staged macro-financial events, joint multi-asset anchors,
  deterministic regeneration, and time-gated disclosure of the shared
  20-series panel.
\item \textbf{Process-aware evaluation.} The benchmark records persistent
  research state, portfolio proposals, executed holdings, factor
  lifecycle events, and failure states, allowing evaluation to
  distinguish outcome quality from internal degeneration or execution
  inconsistency.
\item \textbf{Controlled evaluation contract.} FORESIGHT-9 standardizes the
  market panel, portfolio constraints, cadence, costs, temporal progression,
  and account-write/gating interface while preserving each framework's native
  research, proposal-generation, and persistent-state mechanisms.
\end{itemize}

We instantiate the benchmark with two agent
frameworks---\textbf{AlphaCrafter} (AC), a multi-agent
miner--screener--trader system, and \textbf{FactorMiner} (FM), an
adaptive factor-research agent---each under two foundation-model
backbones: gpt-5.6-terra and deepseek-v4-flash (``DS''). All $2\times2\times9=36$ runs completed to
the 2035-12-31 horizon from identical initial capital.

To separate online adaptation from simpler sources of portfolio performance,
we use a deterministic reference ladder. Buy-and-hold EW isolates passive
exposure by allocating $1/15$ at the boundary and never rebalancing. EW-15
adds only mechanical constant-mix rebalancing on the common ten-trading-day
grid. Inverse-volatility (60d) adds a causal trailing-risk signal but no
scenario interpretation, factor discovery, or learned state. The adaptive
agents therefore sit above progressively stronger non-learning controls under
the same universe, boundary capital, and cost contract.

The evidence follows the same progression. The retrospective ranking has
median Spearman $\rho=-0.80$ against the nine prospective rankings, and its
winner remains first only once. Prospectively, periodic equal weighting and
inverse volatility each outperform 31 of 36 agent runs, so positive adaptive
returns are not by themselves incremental. Finally, WL8 shows that a run can
keep earning after its research library has collapsed and its executed
portfolio has converged toward a fallback (\S\ref{sec:results}).

FORESIGHT-9 releases the nine worldline panels and narratives, all 36
processed trajectories (NAV, holdings, per-decision factor usage,
factor-library audit trails with admission/eviction reasons, and ensemble
histories), the staged factor inventory (Appendix~\ref{app:factors}),
deterministic references, frozen RR inputs, and package-relative scripts for
verification and regeneration of the reported quantitative summaries.

%% file: sections/related_work.tex
\section{Related Work}
\label{sec:related}

\paragraph{Retrospective evaluation of financial agents is not enough.}
LLM-agent systems now mine alpha factors, screen signals, and manage
portfolios \citep{yu2024finmem,zhang2024finagent,xiao2024tradingagents,
yuan2026alphacrafter,wang2026factorminer}, but they are typically
evaluated by a single backtest on one realized path---exposed to
look-ahead leakage and path overfitting
\citep{glasserman2023lookahead,didisheim2025predictable,zhu2026ktdfin}.
KTD-Fin controls memorized knowledge when evaluating agents on realized
financial history; FORESIGHT-9 instead moves the evaluation environment
itself beyond realized history and additionally audits persistent
adaptation and execution state. A single-history leaderboard cannot tell whether a ranking reflects
decision quality or the difficulty of one draw
\citep{white2000reality,bailey2017pbo}. Existing financial-agent
evaluations therefore provide rich agent mechanisms but limited
variation in the evaluation environment. Unlike an OOS backtest, our
retrospective replay is deliberately in-sample and tests whether the
ranking induced by realized-history evaluation agrees with rankings under
unseen counterfactual futures.

\paragraph{Synthetic paths and predefined stresses fall short as
evaluation environments.}
Stress testing evaluates policies across synthetic adverse scenarios
\citep{glasserman2015stress}; generative time-series models synthesize
alternative trajectories from learned historical dynamics
\citep{yoon2019timegan,wiese2020quantgans}; SynthFin provides a fully
synthetic, time-gated sequential trading benchmark for LLM agents
\citep{synthfin2026}; counterfactual evaluation
estimates off-policy performance from logged data
\citep{dudik2011doubly,thomas2016offpolicy}. These approaches establish complementary synthetic and counterfactual
evaluation settings. FORESIGHT-9 specifically targets persistent,
time-gated environments for adaptive foundation-model
agents: the agent under test lives inside the environment for years of
in-world time, must maintain state across stages, and must never observe
beyond the current in-world date. FORESIGHT-9's worldlines are
documented counterfactual continuations of a shared history designed
precisely for that use.

\paragraph{Process-aware evaluation of long-running agents.}
Agent benchmarks moved evaluation from static datasets to interactive
environments
\citep{liu2024agentbench,zhou2024webarena,xie2024osworld}, with
follow-up work pushing trajectory-level scoring---intermediate states,
tool-use quality, latency, cost
\citep{ma2024agentboard,yao2025taubench,lu2025toolsandbox}---and process
supervision distinguishing intermediate-step feedback from outcome-only
supervision \citep{uesato2022process,lightman2024verify}. FORESIGHT-9
combines forward scenario variation with persistent process telemetry in
a financial decision environment: admission and eviction audits are
step-level records of research decisions, and the run-evidence layer
captures operational failure and recovery as first-class evaluation
objects.

%% file: sections/worldlines.tex
\section{The FORESIGHT-9 Benchmark}
\label{sec:worldlines}

The evaluation unit of FORESIGHT-9 is a worldline (WL): a complete
forward market path with a documented anchor table, staged scenario
transitions, a volatility profile, and fixed within-stage lead waypoints. The nine WLs
are controlled stress scenarios, not forecasts: the agent-elicited 10-year
plausibility ranges from the scenario debate (\S\ref{sec:worldlines-origin}) are
provenance metadata and never enter evaluation as sampling frequencies or
likelihood weights.

\subsection{Shared information boundary and panel contract}
\label{sec:worldlines-panel}

All nine WLs fork from a shared information boundary
$t_0=\text{2026-07-15}$. Before $t_0$, every run sees the same daily
panel of 20 series: 15 portfolio market exposures---equity indices $\{$SPX,
NDX, SOX, SX5E, N225, HSI, 000300.SH, 000688.SH$\}$, commodities $\{$XAU,
COPPER, WTI$\}$, crypto $\{$BTC, ETH$\}$, and sovereign yields $\{$US10Y,
CN10Y$\}$---plus 5 observation-only signals (DXY, USDCNY, USDJPY, EURUSD,
VIX). Sovereign yields are quoted in level points and traded directly
as synthetic yield-level exposures---a long position profits when the
yield level rises---directionally analogous to a short-duration
exposure, though position P\&L is the percentage change of the level
rather than duration-scaled (Appendix~\ref{app:worldlines}). Returns on all exposures are combined
in local-currency percentage terms without FX conversion, so a portfolio
is a normalized cross-market allocation rather than a
currency-denominated account. Agent policy state is frozen through the close of 2026-07-15. All runs share
the observed 2026-07-16 execution/market anchor, after which the synthetic
worldlines diverge from 2026-07-17 through 2035-12-31 ($\sim$2{,}468 daily
marks). An agent at in-world date $t$ sees exactly
the rows with date $\le t$; any exception is a failed data audit, not a
modeling choice. Panel construction details and data-quality boundaries
are documented in Appendix~\ref{app:worldlines}.

\subsection{Scenario origination: a specialist agent team and debate}
\label{sec:worldlines-origin}

The scenario set was produced by a three-role financial-research agent
team covering strategy (policy), macro-finance, and chief analyst
(synthesis), run on a commercial financial-data and agent platform. Each
role proposed three 2026--2035 scenarios, followed by three rounds of
cross-review (independent derivation, bidirectional cross-review,
defended responses), scoring every scenario on logical rigor, an agent-elicited
10-year plausibility range, and shock severity. The resulting scenario
specifications---staged narratives and joint multi-asset endpoints---are
released as provenance metadata attached to the numeric stage dates and
endpoints consumed by the trajectory generator (\S\ref{sec:worldlines-generator}). The full score matrix is reported in Appendix~\ref{app:provenance}, and the
anchor-level provenance is documented in Appendix~\ref{app:worldlines}.

\subsection{Generator: scenario anchors with bridge noise}
\label{sec:worldlines-generator}

Each WL is specified as a sequence of 4--5 \emph{stages}, each with a
named narrative endpoint and a terminal date
(Table~\ref{tab:families}; full stage narratives in
Appendix~\ref{app:worldlines}). Prices are interpolated in log space
between stage endpoints; yield and volatility paths follow the stage
specifications; foreign-exchange series without an explicit trajectory
are derived from DXY via declared betas. Within each stage, local
variation is generated by a seeded Brownian bridge
\citep{karatzas1991brownian} while preserving the scenario-defined
endpoints,
\begin{equation}
    \ell_u = \tfrac{T-u}{T}\,\ell_0 + \tfrac{u}{T}\,\ell_T
      + \sigma\!\left(B_u-\tfrac{u}{T}B_T\right),
    \qquad u=0,\ldots,T,
    \label{eq:gbb}
\end{equation}
whose volatility $\sigma$ is calibrated to the realized volatility of
the warmup window and whose endpoints are pinned to zero, so the
perturbation hits each designated stage endpoint exactly: scenario
anchors determine the macro path, and the bridge supplies only local
variation. At stage boundaries the path is \emph{re-anchored} to the
next stage's endpoint table, preventing accumulated noise from drifting
a WL away from its macro scenario. A fixed within-stage lead waypoint controls
partial movement toward each endpoint; it is a numeric construction rule, not
an agent news channel. The generator records the versioned seed contract and
parameters of every stage, making each WL deterministic conditional on its manifest.
The local bridge perturbations are deterministic scenario texture rather than
a fitted joint return model: their cross-asset covariance is not calibrated to
an empirical covariance matrix beyond the shared anchor construction.

\begin{table}[t]
\centering
\small
\setlength{\tabcolsep}{4pt}
\begin{tabular}{@{}llp{5.4cm}@{}}
\toprule
WL & Worldline crisis & Main transmission \\
\midrule
WL1 & Geopolitical & conflict $\to$ sanctions $\to$ trade fragmentation \\
WL2 & Technology fragmentation & internet splintering $\to$ bipolar digital economy \\
WL3 & Nuclear escalation & nuclear test $\to$ conventional conflict $\to$ new nuclear order \\
WL4 & Sovereign debt / liquidity & Treasury stress $\to$ Minsky moment $\to$ stagflation \\
WL5 & Currency disorder & JPY disorderly devaluation $\to$ Asian currency fragmentation \\
WL6 & Inflation regime & inflation re-ignition $\to$ credibility collapse $\to$ stagflation \\
WL7 & Financial-system fragility & fragility build-up $\to$ algorithmic bank run $\to$ regulatory rebuild \\
WL8 & Energy & Hormuz blockade $\to$ energy-system collapse $\to$ restructuring \\
WL9 & Pandemic / automation & labor shock $\to$ recession $\to$ automation supercycle \\
\bottomrule
\end{tabular}
\caption{\textbf{The nine worldlines at a glance.} Each worldline is a
sequence of 4--5 stages; the full stage end dates and narratives are in
Table~\ref{tab:stages} (Appendix~\ref{app:worldlines}). All WLs end
2035-12-31.}
\label{tab:families}
\end{table}

\subsection{Auditability}
\label{sec:worldlines-audit}

Three properties make a WL auditable rather than merely synthetic:
(i)~\emph{endpoint anchoring}---the stage endpoint table is part of the
released manifest, so any realized path can be checked against its
declared scenario; (ii)~\emph{leakage control}---the information boundary
is enforced by the panel contract of \S\ref{sec:worldlines-panel}, and fixed
lead waypoints are numeric path parameters rather than disclosed events;
(iii)~\emph{determinism}
---conditional on the manifest and seed, regeneration is reproducible,
and the market panel consumed by the four configurations within each worldline
is byte-identical.
All nine proposed scenarios were retained; no worldline was selected or removed
using agent performance. We
also record what the WLs are \emph{not}: they are not sampled from a
calibrated real-world probability law, and robustness summaries across
WLs (medians, quantiles, worst-WL) describe behavior under documented
scenario spread, not probabilistic expectations.

%% file: sections/agents.tex
\section{Evaluated Agents and Protocol}
\label{sec:agents}

FORESIGHT-9 evaluates agents as test subjects: they observe the panel,
maintain persistent state, and emit portfolios through a fixed execution
interface, while their internal search and self-evolution mechanisms
remain intact. We evaluate two frameworks---FactorMiner (FM) and
AlphaCrafter (AC)---under two foundation-model backbones: gpt-5.6-terra and
deepseek-v4-flash (``DS'').

\subsection{FactorMiner (FM): adaptive factor research}
\label{sec:agents-fm}

FM follows the FactorMiner framework \citep{wang2026factorminer} and runs
a mining--admission--combination loop with a strict separation of
concerns: only \emph{mining} consumes the LLM; admission, combination, and
forward execution are deterministic. In the shared warmup
(2020-01-01--2026-07-15) the loop mines a large candidate pool; online,
each 10-trading-day window appends one further iteration. Admission is a
deterministic contract built on one principle: at a fifteen-asset
cross-section, no single statistic can be trusted to certify
significance, so the gate values---$|\mathrm{IC}|\ge 0.007$ and
$|\mathrm{ICIR}|\ge 0.084$ on the window mean daily series---are fixed by
stipulation and deliberately permissive, and selectivity is engineered
downstream where it can be audited stage by stage
(Figure~\ref{fig:admission}): Gate 1 removes candidates without
measurable signal, Gate 2 removes unstable ones, and Gate 3 enforces
diversity by routing each survivor on its conflict count $k$ against the
incumbents, so that a correlated but strictly better candidate can
displace a weaker incumbent instead of being silently blocked. Capacity
then keeps the library at its thirty best slots by
$q_i=|\mathrm{IC}_i|\cdot|\mathrm{ICIR}_i|$---a greedy,
diversity-constrained library update rule, repeatedly re-optimized,
pairwise near-orthogonal ($\rho<0.5$), and trim-protected against stale
members. At each decision the combination step takes the IC-weighted
top-10 of the current library, preserving
$\operatorname{sign}(\mathrm{IC})$; absent a valid ensemble the
contract's built-in fallback is the equal-weight $1/15$ vector. Every
lifecycle event---proposed, parsed, screened, rejected, admitted,
evicted---is appended with the candidate's formula to an auditable
factor-lifecycle log.

The complete admission funnel is shown in Figure~\ref{fig:admission} in Appendix~\ref{app:factors}.

\subsection{AlphaCrafter (AC): multi-agent miner--screener--trader}
\label{sec:agents-ac}

AC follows the AlphaCrafter framework \citep{yuan2026alphacrafter} and
chains specialist agents inside a per-WL sandbox workspace, each research
cycle comprising three parallel Miners, one Screener, and one Trader. The
shared warmup runs 40 such cycles on the pre-boundary history, and each WL
run is seeded from its own experiment's (strictly disjoint) warmup
workspace. Miner agents each propose a research idea, implement and run a
metrics stage, and persist a factor artifact; admission recomputes
pairwise correlations on the artifacts' actual signals rather than
trusting the miner's self-report, then applies the same stipulated
IC/ICIR gates and the same conflict-count routing as FM ($k\ge2$ reject;
$k=1$ quality duel on $q=|\mathrm{IC}|\cdot|\mathrm{ICIR}|$; $k=0$
admit), with the capacity-30 trim evicting the lowest-$q$ member and
every eviction carrying a machine-readable reason. The screener selects
the active ensemble ($\le$10 factors, quality/IC-tilted weights); the
trader consumes the ensemble, computes asset preferences, and emits
target weights. AC's native path has no DSL parser---factor generation is
proposal-and-screen rather than expression search---and we make no claim
that AC and FM perform identical factor search. Account state, the
per-cycle factor-library audit trail, research memory, and the evolving
workspace are all released artifacts.

\subsection{Benchmark contract and evaluation setup}
\label{sec:agents-adapt}

Both frameworks were wired to the benchmark through a thin, logged
adaptation layer; nothing inside either agent's search or evolution loop
was rewritten.

\begin{itemize}
\item \textbf{Unified panel and universe.} Both frameworks consume the
  same 15+5 series with the same naming and calendar; within each worldline,
the corresponding market panel is byte-identical across the four configurations.
\item \textbf{Decision cadence and the proposal/gate contract.} Both
  frameworks decide on the same 10-trading-day grid. A decision may only
  produce a \emph{proposal}; a deterministic execution layer is the sole
  writer to accounts. With $w$ the current and $w^\ast$ the proposed
  target weights and $\hat r$ the agent's forecast returns over the
  10-day horizon,
  \begin{equation*}
    \tau = \tfrac12\textstyle\sum_k |w^\ast_k - w_k|, \quad
    e = 10^4 \textstyle\sum_k (w^\ast_k - w_k)\,\hat r_k,
  \end{equation*}
  and the rebalance executes only if the run is the initial full
  allocation or $e > 3\tau$; the realized cost is
  $3\,\mathrm{bps}\times\tau\times V$. A gated-out decision persists the
  proposal and forecasts but leaves holdings untouched---a logged
  no-trade, not a failure. Observations are masked through the decision-date
  close; framework-specific fill conventions and the quantified AC same-day
  overlap are documented in Appendix~\ref{app:metrics}. The simulator executes
  emitted weights verbatim, so any proposal/execution divergence is auditable.
\item \textbf{Portfolio constraints and costs.} Long-only, fully
  invested, non-negative weights summing to 1 with zero cash; first
  allocation free; 3\,bps one-way on migrated notional at every
  rebalance.
\item \textbf{Per-WL isolation and persistence.} Each run lives in an
  isolated sandbox with persisted state and supports checkpoint/resume
  without changing in-world time.
\item \textbf{Failure handling and temporal progression.} Each cycle advances
  the evaluation clock by exactly 10 trading days regardless of module
  failures; when no executable proposal is produced, current holdings are
  marked to market and the event is logged.
\item \textbf{Trajectory logging and process telemetry.} We record observations,
  tool calls, factor-validation outcomes, proposals, executions, and skip
  reasons; reflection and bounded-memory timestamps are retained in the
  process audit trail, while hidden reasoning text is not an evaluation target.
\end{itemize}

This protocol yields 36 completed runs; per-run metrics, NAV conventions, and
model configurations are reported in Appendices~\ref{app:metrics} and
\ref{app:models}.

%% file: sections/results.tex
\FloatBarrier
\section{Results}
\label{sec:results}

We report 36 completed prospective runs, Native-Policy Retrospective Replay
(RR), and three deterministic references; replay and execution conventions are
detailed in Appendix~\ref{app:rr}.

\subsection{Retrospective Rankings Do Not Reliably Transfer}
\label{sec:results-rr}

We first ask whether the ranking induced by realized-history evaluation
predicts prospective adaptive performance. For each configuration, we freeze
the learned policy at the shared 2026-07-15 boundary and execute its
framework-specific native policy over the realized historical panel; all
resulting proposals are evaluated under the common causal replay contract. RR is deliberately
retrospective and in-sample: the final learned policy is replayed on the same
history from which it was obtained. We therefore do not interpret RR returns
as estimates of future performance; the question is only whether its ranking
agrees with rankings under unseen worldlines. Full reconstruction details are
reported in Appendix~\ref{app:rr}.

\begin{table}[H]
\centering
\small
\setlength{\tabcolsep}{4pt}
\begin{tabular}{lrrrr}
\toprule
Configuration & RR ann. return & RR rank & Prospective median rank & WLs ranked 1st \\
\midrule
FM--DS & 21.64\% & 1 & 4 & 1/9 \\
AC--terra & 18.81\% & 2 & 3 & 0/9 \\
FM--terra & 17.19\% & 3 & 1 & 5/9 \\
AC--DS & 16.09\% & 4 & 2 & 3/9 \\
\bottomrule
\end{tabular}
\caption{\textbf{Retrospective-to-prospective rank transfer.}
RR ranks configurations by terminal cumulative return within the realized-history
replay (the annualized return is shown only as a historical-performance anchor);
the prospective columns summarize agent-only ranks across WL1--WL9.
Full RR portfolio metrics are reported in Appendix~\ref{app:rr}.}
\label{tab:rr-summary}
\end{table}

The RR ordering is defined by terminal cumulative return within the replay;
AC--terra has the highest RR Sharpe (1.496), providing a useful metric
sensitivity check reported in Appendix~\ref{app:rr}. FM--DS falls from first retrospectively to prospective median
rank 4, whereas FM--terra moves from third to median rank 1. The reordering
extends beyond these medians. Across the nine worldlines, the RR ordering has
a mean Spearman correlation of $-0.42$ (median $-0.80$) with the prospective
ranking, and only 17 of 54 pairwise orderings (31.5\%) are preserved. The
correspondence varies substantially across worldlines, ranging from complete reversal on WL8
($\rho=-1.0$) and a near-reversal on WL1 ($\rho=-0.8$)
to strong agreement on WL6 ($\rho=+0.8$). These are descriptive
statistics over four agent configurations; no significance test or $p$-value
is claimed.

\begin{figure}[t]
\centering
\includegraphics[width=\columnwidth]{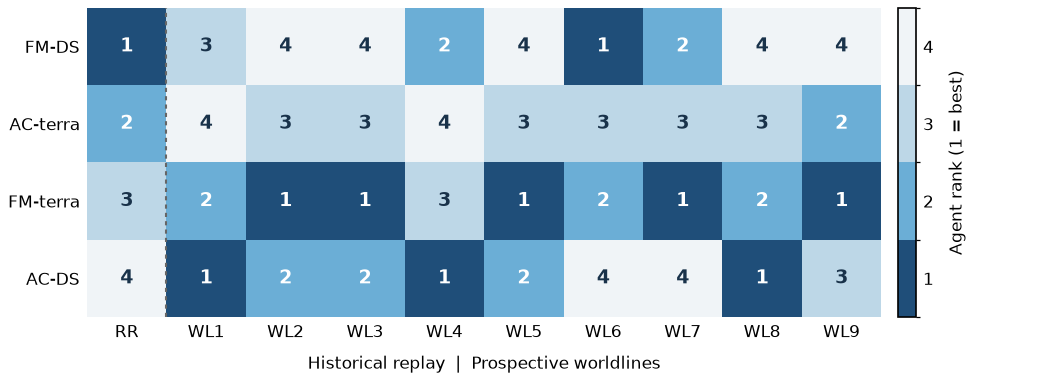}
\caption{\textbf{Retrospective-to-prospective rank transfer.} Each cell shows
the agent-only rank (1 = best) under Native-Policy RR or within a prospective
worldline. The realized-history ordering transfers inconsistently across
worldlines (mean Spearman $\bar\rho=-0.42$, median $\rho=-0.80$); the RR
leader FM--DS remains first on only one of nine worldlines.}
\label{fig:rank-transfer}
\end{figure}

\subsection{Outcome Robustness Across Worldlines}
\label{sec:results-outcome}

\begin{figure}[t]
\centering
\includegraphics[width=0.92\columnwidth]{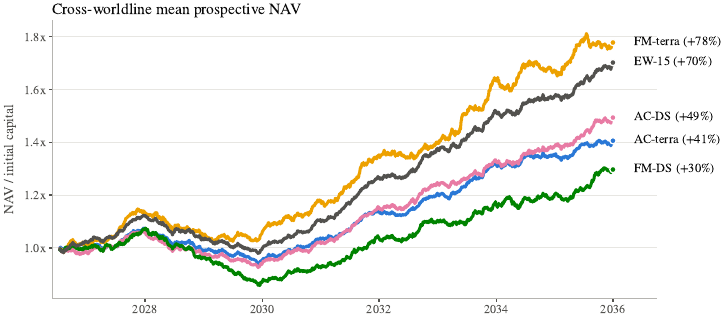}
\caption{\textbf{Cross-worldline mean normalized prospective NAV.} The four
agent--backbone configurations are compared with the periodically rebalanced
equal-weight reference (EW-15). At each in-world date, each run is normalized
by the common boundary capital and averaged with equal descriptive weight
across WL1--WL9; the curve is a cross-scenario summary, not a
probability-weighted expected future path. Deterministic baselines are
reported separately in Appendix~\ref{app:nav}.}
\label{fig:mean-nav}
\end{figure}

Performance varies substantially across both worldlines and model
backbones. FM has the highest mean return under terra ($+77.8\%$ versus
$+40.6\%$ for AC), whereas AC leads under DS ($+49.4\%$ versus $+29.7\%$).
The inversion is not driven by one scenario: FM--terra ranks first among
agents on five of nine worldlines, whereas FM--DS does so only once. Individual
worldlines can reverse the ordering sharply---for example, WL5 yields $+138.7\%$
for FM--terra but only $+8.4\%$ for FM--DS. Full per-worldline cumulative
returns are reported in Appendix~\ref{app:metrics}.

The prospective matrix therefore substantially reorders the realized-history
leaderboard above. Full per-worldline trajectories and metrics appear in
Appendices~\ref{app:nav} and \ref{app:metrics}.

\subsection{Comparison with Deterministic References}
\label{sec:results-ew}

The reference ladder separates three simpler sources of return. Passive
buy-and-hold EW has a mean cumulative return of $+22.0\%$ and beats 3/36
agent runs. Adding only ten-day constant-mix rebalancing raises EW-15 to
$+70.2\%$ and 31/36 paired wins. A causal inverse-volatility rule reaches
$+70.4\%$ and the same 31/36 wins without scenario interpretation, factor
discovery, or learned state (Appendix~\ref{app:metrics},
Table~\ref{tab:det-baselines}). Mechanical rebalancing and simple risk
allocation therefore capture much of the attainable prospective performance;
positive agent return alone cannot be attributed to adaptive intelligence.

The mismatch also appears relative to EW-15. AC--terra, AC--DS, and FM--DS
beat it on 0/9, 1/9, and 0/9 worldlines, while FM--terra does so on 4/9.
FM--DS is the only
configuration with positive retrospective annualized excess (+2.46 pp), yet
it beats EW-15 on 0/9 prospective worldlines. FM--terra shows the opposite
pattern, trailing EW retrospectively but beating it on 4/9 worldlines. We
call these differences ``apparent excess return,'' not alpha.
Window-level persistence diagnostics are reported in Appendix~\ref{app:significance}.

\subsection{Process Consistency}
\label{sec:results-process}

A favorable terminal return does not imply that the agent's intended
adaptation mechanism remained functional. RR diagnoses instability in the
realized-history ranking; process traces ask a different question: whether a
favorable return was still produced by the intended adaptive mechanism.
Because the benchmark's promise is \emph{self-evolution}, we evaluate not
only outcomes but whether the three states a run maintains stay coherent:
the \emph{research state} (the live factor library), the \emph{declared
decision state} (the ensemble and portfolio the agent records), and the
\emph{executed state} (the holdings actually held). All three are recorded
persistently as trajectory series and terminal registers
(Appendix~\ref{app:process}), so a run is auditable as a decision system
rather than as a terminal score.

In WL8, the live research state collapses while the declared ensemble persists
and the executed portfolio converges toward equal weight; the run can therefore
look successful at the outcome level even after meaningful self-evolution has
ceased. Concretely, AC--DS ends at $+86.3\%$ even though its live factor
library reaches zero and the three persisted states no longer agree.
Appendix~\ref{app:process} tabulates the lifecycle registers for all runs.

Similar terminal snapshots conceal heterogeneous audit trails, from stable
libraries to churn ending in zero survivors. Across runs, transaction costs are
modest (at most 126\,bps of initial capital), so they do not explain the return
dispersion.

%% file: sections/limitations.tex
\section{Limitations}
\label{sec:limitations}

FORESIGHT-9 uses one seed per configuration and worldline; multi-seed
replication would separate agent variance from worldline variance but would
substantially increase the compute budget. The nine worldlines are controlled
scenarios rather than probabilistic forecasts: no worldline is assigned a
likelihood, and cross-worldline summaries describe the documented scenario
spread rather than an expectation under a calibrated return law. Although the
anchors enforce coherent multi-asset endpoints, the intervening trajectories
are not claimed to be samples from a calibrated joint distribution, and their
plausibility has not been independently validated by domain experts. The
large gap between periodically rebalanced EW-15 and passive buy-and-hold should
be interpreted within this benchmark: staged, repeatedly re-anchored path
construction may favor constant-mix rebalancing, so its magnitude is not an
estimate of a general real-market rebalancing premium.

Operational events such as service interruptions and retries do not enter
in-world outcomes because the calendar advances deterministically. The
benchmark covers two frameworks and two backbones; its long-only, fully
invested contract evaluates relative cross-market allocation rather than cash
timing or short selling.
Forward auxiliary fields are deterministic close-derived proxies rather than
independent market-microstructure channels; agents that rely on volume, VWAP,
or intraday structure are therefore evaluated under a restricted signal
environment.

Native-Policy Retrospective Replay (RR) is deliberately in-sample and freezes
the learned boundary policy rather than replaying the complete historical
learning trajectory. It preserves framework-specific learned policies but
standardizes replay execution, so it tests whether retrospective scoring of
the final learned state predicts prospective adaptive performance rather than
reproducing each framework's original end-to-end execution semantics. RR does
not identify historical contamination as a causal mechanism. Rank-transfer
statistics involve only four configurations and are reported descriptively
without significance testing; discrepancies cannot be attributed uniquely to
regime shift or historical familiarity.

%% file: sections/conclusion.tex
\section{Conclusion}
\label{sec:conclusion}

FORESIGHT-9 exposes three successive failures of outcome-only evaluation.
First, a learned-policy ranking from realized history transfers poorly to
alternative futures: median Spearman correlation is $-0.80$, and the RR
winner remains first in only $1/9$ worldlines. Second, prospective profit does
not establish incremental value from complexity: periodic equal weighting and
inverse volatility each outperform 31/36 agent--worldline runs. The third failure is
mechanistic: an agent can continue earning returns after it has effectively
stopped adapting. In WL8, the live research library collapses while the
declared ensemble persists and executed holdings converge toward equal weight.

Reliable agent evaluation therefore requires future robustness, incremental
value over simple controls, and functional adaptation. FORESIGHT-9 combines
retrospective rank-transfer diagnosis, controlled alternative futures, a
deterministic reference ladder, and concrete audits of research, declared,
and executed state to test these requirements together.

%% file: sections/appendix_nav.tex
\clearpage
\section{Additional NAV Views}
\label{app:nav}
\begin{figure}[H]
\centering
\includegraphics[width=0.85\textwidth]{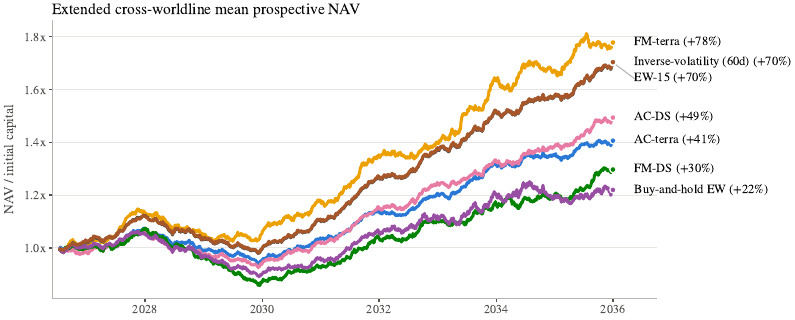}
\caption{\textbf{Extended cross-worldline mean normalized prospective NAV.} The four
adaptive configurations and three deterministic references (EW-15,
buy-and-hold EW, and inverse-volatility 60d) are
shown on a common boundary-capital normalization. This aggregate view
averages the normalized daily NAV pointwise across the nine fixed worldlines
with equal descriptive weight; it is a cross-scenario summary rather than a
probability-weighted or realizable future path. The worldline-level baseline
heterogeneity is shown separately below.}
\label{fig:mean-nav-all}
\end{figure}

\begin{figure}[H]
\centering
\includegraphics[width=0.85\textwidth]{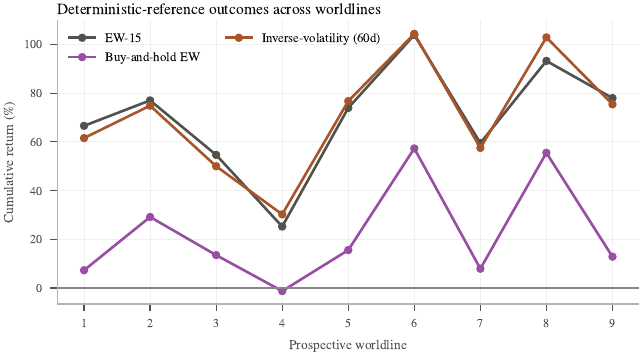}
\caption{\textbf{Deterministic baseline comparison across prospective
worldlines.} The figure isolates EW-15, buy-and-hold EW, and inverse-volatility
(60d); cumulative returns are computed under the same shared information boundary, 2026-07-16 observed anchor, worldline
panels, and fee convention as the agent runs.}
\label{fig:baseline-comparison}
\end{figure}

\begin{figure}[H]
\centering
\includegraphics[width=0.85\textwidth]{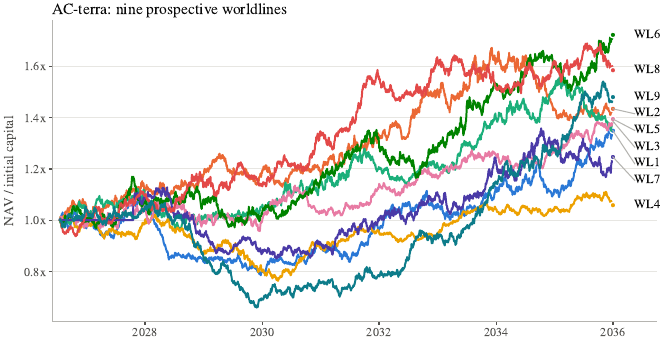}
\caption{\textbf{AC--terra across the nine worldlines} (nine solid runs;
colorblind-validated palette; right-edge labels mark the terminal
worldline). Conventions are identical in
Figures~\ref{fig:overlay-ac-ds}--\ref{fig:appwl9}.}
\label{fig:overlay-ac-terra}
\end{figure}

\begin{figure}[H]
\centering
\includegraphics[width=0.85\textwidth]{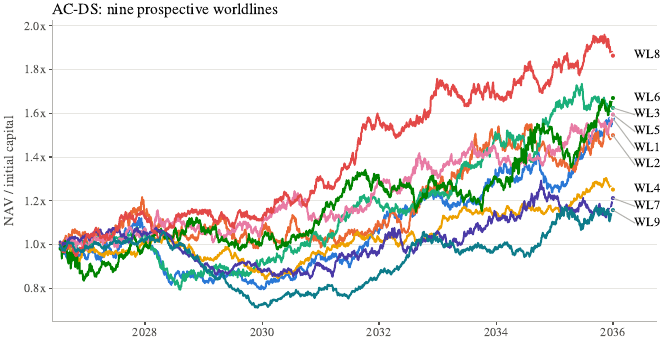}
\caption{\textbf{AC--DS across the nine worldlines.}}
\label{fig:overlay-ac-ds}
\end{figure}

\begin{figure}[H]
\centering
\includegraphics[width=0.85\textwidth]{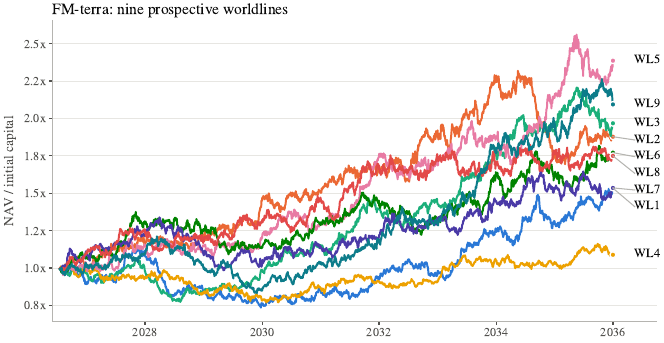}
\caption{\textbf{FM--terra across the nine worldlines.}}
\label{fig:overlay-fm-terra}
\end{figure}

\begin{figure}[H]
\centering
\includegraphics[width=0.85\textwidth]{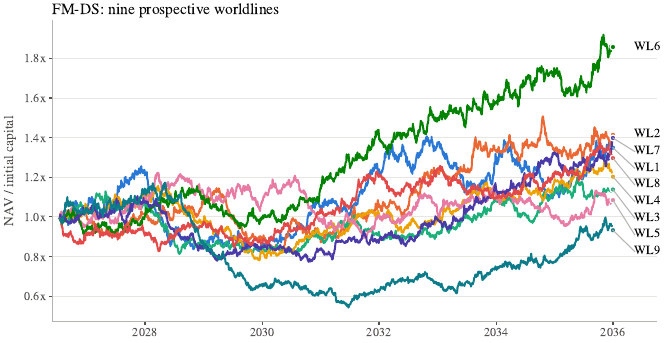}
\caption{\textbf{FM--DS across the nine worldlines.} The framework that
leads under terra (Figure~\ref{fig:overlay-fm-terra}) trails under DS:
the per-worldline ranking is different across the two evaluated backbones.}
\label{fig:overlay-fm-ds}
\end{figure}

Per-version overlays across the nine worldlines
(Figures~\ref{fig:overlay-ac-terra}--\ref{fig:overlay-fm-ds}), followed
by one panel per worldline: the four completed runs (AC--terra, AC--DS,
FM--terra, FM--DS) against the 15-asset equal-weight baseline
(Appendix~\ref{app:metrics} tabulates the corresponding risk-adjusted
metrics). All curves are solid in a validated five-color scheme; end-of-horizon
labels give terminal returns. AC has no released raw daily account-NAV series.
Its displayed daily curves are reconstructions: each segment marks the recorded
post-trade holdings on daily closes and uses a log-space multiplicative bridge
to hit the next recorded pre-trade NAV exactly; recorded cost then gives the
post-trade anchor. The final bridge hits canonical terminal net assets. Across
18 curves, event-anchor relative error is at most $2.3\times10^{-16}$ and the
absolute terminal correction is below $1.5\%$ (mean $0.49\%$, maximum
$1.23\%$). Unrecorded ten-day marks are reconstructed values, not raw account
observations.

\begin{figure}[H]
\centering
\includegraphics[width=0.85\textwidth]{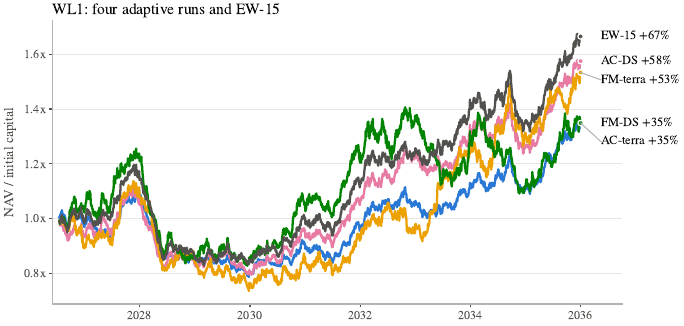}
\caption{\textbf{WL1: four runs vs.\ the equal-weight baseline.} AC--terra (blue), AC--DS (pink), FM--terra (amber), FM--DS (green), EW-15 (gray); labels give terminal returns.}
\label{fig:appwl1}
\end{figure}

\begin{figure}[H]
\centering
\includegraphics[width=0.85\textwidth]{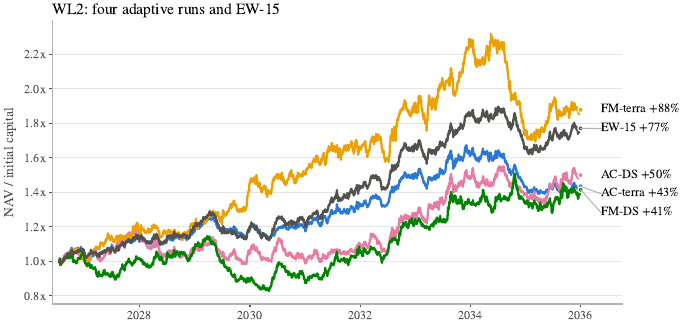}
\caption{\textbf{WL2: four runs vs.\ the equal-weight baseline.} AC--terra (blue), AC--DS (pink), FM--terra (amber), FM--DS (green), EW-15 (gray); labels give terminal returns.}
\label{fig:appwl2}
\end{figure}

\begin{figure}[H]
\centering
\includegraphics[width=0.85\textwidth]{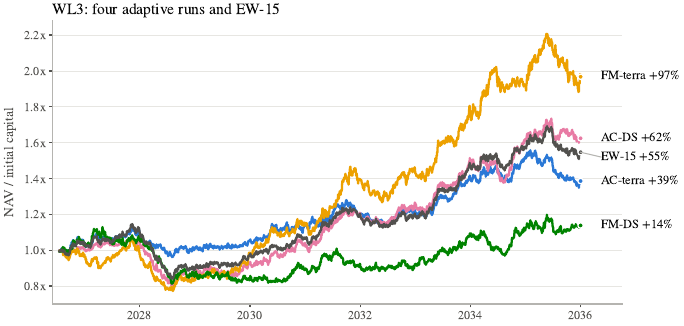}
\caption{\textbf{WL3: four runs vs.\ the equal-weight baseline.} AC--terra (blue), AC--DS (pink), FM--terra (amber), FM--DS (green), EW-15 (gray); labels give terminal returns.}
\label{fig:appwl3}
\end{figure}

\begin{figure}[H]
\centering
\includegraphics[width=0.85\textwidth]{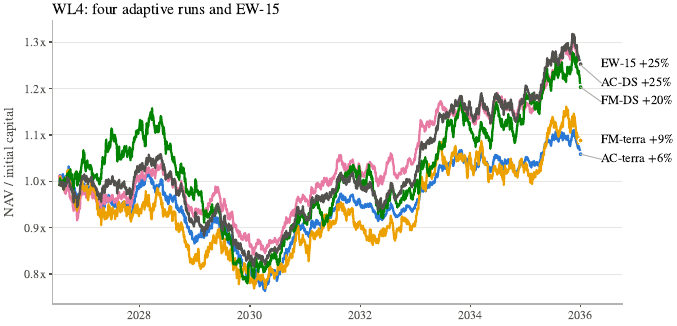}
\caption{\textbf{WL4: four runs vs.\ the equal-weight baseline.} AC--terra (blue), AC--DS (pink), FM--terra (amber), FM--DS (green), EW-15 (gray); labels give terminal returns.}
\label{fig:appwl4}
\end{figure}

\begin{figure}[H]
\centering
\includegraphics[width=0.85\textwidth]{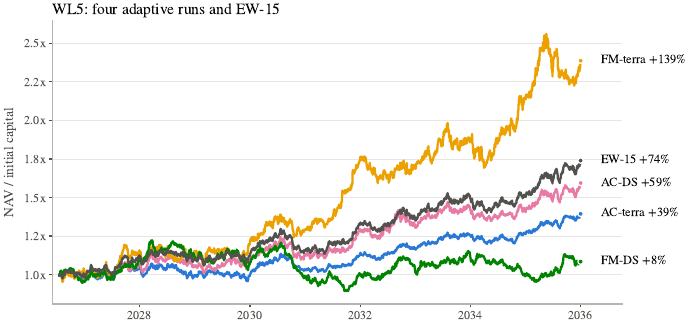}
\caption{\textbf{WL5: four runs vs.\ the equal-weight baseline.} AC--terra (blue), AC--DS (pink), FM--terra (amber), FM--DS (green), EW-15 (gray); labels give terminal returns.}
\label{fig:appwl5}
\end{figure}

\begin{figure}[H]
\centering
\includegraphics[width=0.85\textwidth]{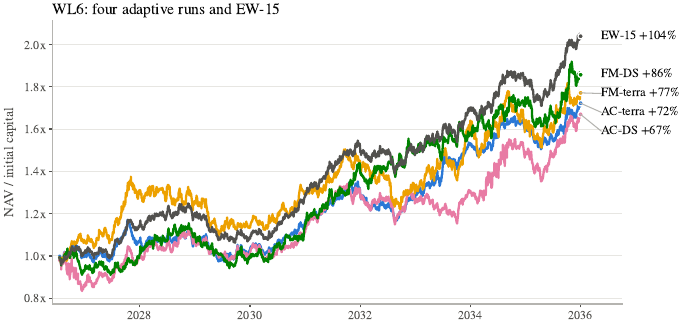}
\caption{\textbf{WL6: four runs vs.\ the equal-weight baseline.} AC--terra (blue), AC--DS (pink), FM--terra (amber), FM--DS (green), EW-15 (gray); labels give terminal returns.}
\label{fig:appwl6}
\end{figure}

\begin{figure}[H]
\centering
\includegraphics[width=0.85\textwidth]{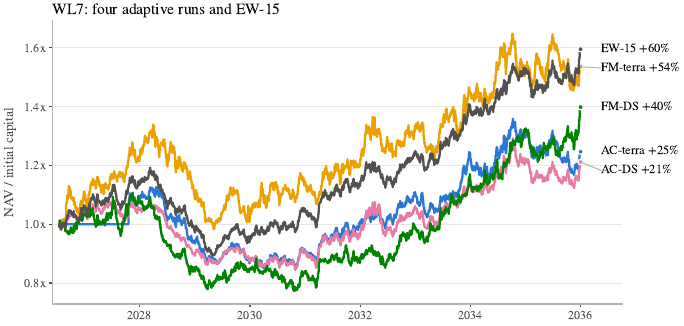}
\caption{\textbf{WL7: four runs vs.\ the equal-weight baseline.} AC--terra (blue), AC--DS (pink), FM--terra (amber), FM--DS (green), EW-15 (gray); labels give terminal returns.}
\label{fig:appwl7}
\end{figure}

\begin{figure}[H]
\centering
\includegraphics[width=0.85\textwidth]{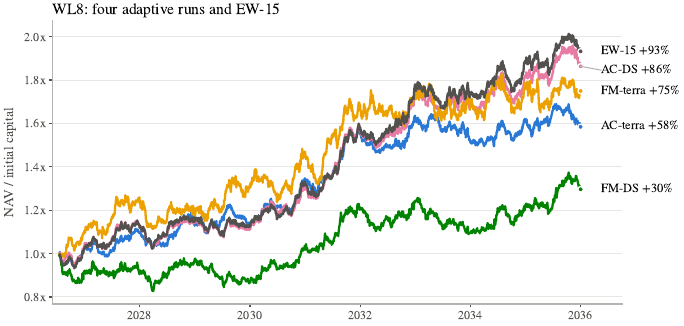}
\caption{\textbf{WL8: four runs vs.\ the equal-weight baseline.} AC--terra (blue), AC--DS (pink), FM--terra (amber), FM--DS (green), EW-15 (gray); labels give terminal returns.}
\label{fig:appwl8}
\end{figure}

\begin{figure}[H]
\centering
\includegraphics[width=0.85\textwidth]{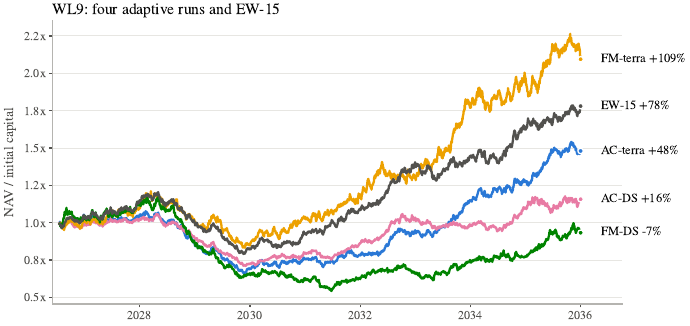}
\caption{\textbf{WL9: four runs vs.\ the equal-weight baseline.} AC--terra (blue), AC--DS (pink), FM--terra (amber), FM--DS (green), EW-15 (gray); labels give terminal returns.}
\label{fig:appwl9}
\end{figure}

%% file: sections/appendix_metrics.tex
\clearpage
\section{Per-Run Risk-Adjusted Metrics}
\label{app:metrics}
Daily-series metrics per run ($r_f=0$; AC series are the daily reconstructions of Appendix~\ref{app:nav}). Median refers to the across-worldline median of each column.

\subsection{Full Prospective Outcome Matrix}
\label{app:outcome-matrix}

The complete cumulative-return matrix supporting the summary discussion in
Section~\ref{sec:results-outcome} is provided here for auditability.

\begin{table}[H]\centering\small
\begin{tabular}{lrrrrr}
\toprule
Worldline & AC-terra & AC-DS & FM-terra & FM-DS & EW-15 \\
\midrule
WL1 & $+34.9\%$ & $+57.5\%$ & $+53.4\%$ & $+34.9\%$ & $+66.6\%$ \\
WL2 & $+43.4\%$ & $+49.9\%$ & $+87.8\%$ & $+41.4\%$ & $+77.0\%$ \\
WL3 & $+38.6\%$ & $+62.4\%$ & $+96.8\%$ & $+13.8\%$ & $+54.7\%$ \\
WL4 & $+5.8\%$  & $+25.1\%$ & $+8.8\%$  & $+20.3\%$ & $+25.2\%$ \\
WL5 & $+39.4\%$ & $+59.4\%$ & $+138.7\%$& $+8.4\%$  & $+73.9\%$ \\
WL6 & $+72.2\%$ & $+67.0\%$ & $+77.1\%$ & $+85.7\%$ & $+103.9\%$\\
WL7 & $+24.7\%$ & $+21.2\%$ & $+53.5\%$ & $+39.8\%$ & $+59.5\%$ \\
WL8 & $+58.4\%$ & $+86.3\%$ & $+75.0\%$ & $+29.6\%$ & $+93.2\%$ \\
WL9 & $+48.0\%$ & $+15.7\%$ & $+109.4\%$& $-6.7\%$  & $+78.0\%$ \\
\midrule
Mean   & $+40.6\%$ & $+49.4\%$ & $+77.8\%$ & $+29.7\%$ & $+70.2\%$ \\
Median & $+39.4\%$ & $+57.5\%$ & $+77.1\%$ & $+29.6\%$ & $+73.9\%$ \\
Range  & $[+5.8,+72.2]$ & $[+15.7,+86.3]$ & $[+8.8,+138.7]$ & $[-6.7,+85.7]$ & $[+25.2,+103.9]$ \\
\bottomrule
\end{tabular}
\caption{\textbf{Full prospective outcome matrix.} Cumulative total return
per worldline for the four agent--backbone combinations and the equal-weight
baseline. Within each worldline, all four configurations used the same
byte-identical market panel, portfolio constraints, and fee rule.}
\label{tab:outcome}
\end{table}

\begin{table}[H]\centering\footnotesize
\begin{tabular}{lrrrrr}
\toprule
 & AC-terra & AC-DS & FM-terra & FM-DS & EW-15 \\
\midrule
WL1 & 0.41 & 0.52 & 0.37 & 0.27 & 0.59\\
WL2 & 0.53 & 0.41 & 0.61 & 0.35 & 0.78\\
WL3 & 0.46 & 0.58 & 0.67 & 0.14 & 0.60\\
WL4 & 0.10 & 0.39 & 0.10 & 0.23 & 0.35\\
WL5 & 0.50 & 0.55 & 0.81 & 0.09 & 0.72\\
WL6 & 0.67 & 0.50 & 0.54 & 0.64 & 0.92\\
WL7 & 0.28 & 0.26 & 0.41 & 0.35 & 0.62\\
WL8 & 0.66 & 0.81 & 0.54 & 0.28 & 0.89\\
WL9 & 0.36 & 0.18 & 0.70 & -0.06 & 0.70\\
\bottomrule
\end{tabular}
\caption{\textbf{Sharpe ratio per worldline} (daily NAV, annualized, $r_f=0$).}\label{tab:sharpe-wl}
\end{table}

\begin{table}[H]\centering\footnotesize
\begin{tabular}{lrrrrr}
\toprule
 & AC-terra & AC-DS & FM-terra & FM-DS & EW-15 \\
\midrule
WL1 & -28.4\% & -29.0\% & -35.4\% & -34.1\% & -30.9\%\\
WL2 & -16.9\% & -19.1\% & -27.0\% & -27.8\% & -14.6\%\\
WL3 & -14.2\% & -25.7\% & -27.5\% & -28.0\% & -26.7\%\\
WL4 & -26.2\% & -18.6\% & -25.0\% & -32.7\% & -22.9\%\\
WL5 & -10.3\% & -13.0\% & -14.7\% & -26.9\% & -12.5\%\\
WL6 & -16.5\% & -18.1\% & -20.7\% & -18.5\% & -14.9\%\\
WL7 & -24.5\% & -21.6\% & -26.6\% & -30.1\% & -25.0\%\\
WL8 & -9.1\% & -10.3\% & -12.3\% & -16.9\% & -11.3\%\\
WL9 & -38.3\% & -32.8\% & -31.4\% & -53.8\% & -34.4\%\\
\bottomrule
\end{tabular}
\caption{\textbf{Maximum drawdown per worldline} (daily NAV).}\label{tab:mdd-wl}
\end{table}

\paragraph{Deterministic reference definitions.}
We use three deterministic references. EW-15 rebalances to equal $1/15$
weights on the 10-trading-day decision grid. Buy-and-hold EW initializes the
same equal-weight allocation and never rebalances. Inverse-volatility
rebalances on the same grid using weights proportional to inverse trailing-
60-observation return volatility, normalized to full investment. All three
references are long-only and fully invested, use causal price history over the
same 15 portfolio market exposures, receive a free initial allocation, and
incur the same 3-bps one-way migrated-notional cost thereafter.
All three use no cash, leverage, shorting, stop-loss, or volume/liquidity
information. Decisions use only closes through the decision date, fill at the
next open, and are marked at that day's close. The paired-wins column counts
$\sum_{w=1}^{9}\sum_{a=1}^{4}{\bf 1}[R_{b,w}>R_{a,w}]$; agent returns use
$\mathrm{final\_nav}/1{,}000{,}000-1$.

\begin{table}[H]\centering\scriptsize
\setlength{\tabcolsep}{3pt}
\begin{tabular}{lrrrrr}
\toprule
Strategy & Mean return & Median return & Median Sharpe & WLs $>$ EW-15 & Paired wins vs. agents (/36) \\
\midrule
EW-15 & $+70.2\%$ & $+73.9\%$ & 0.70 & --- & 31/36 \\
Buy-and-hold EW & $+22.0\%$ & $+13.5\%$ & 0.13 & 0/9 & 3/36 \\
Inverse-volatility (60d) & $+70.4\%$ & $+74.9\%$ & 0.69 & 4/9 & 31/36 \\
\bottomrule
\end{tabular}
\caption{\textbf{Deterministic reference comparison.} Returns are cumulative
prospective returns summarized across WL1--WL9; Sharpe is the median of the
per-worldline daily-NAV Sharpe ratios. Agent comparator returns use the fixed
initial-capital denominator $V_0=1{,}000{,}000$ (terminal NAV divided by $V_0$,
minus one), matching the paper convention.}
\label{tab:det-baselines}
\end{table}

The released deterministic-baseline matrix has 27 run rows: EW-15 plus two
causal references (buy-and-hold EW and inverse-volatility 60d), each evaluated
on all nine prospective worldlines.

\subsection{Run-status bookkeeping and NAV conventions}
\label{app:metrics-protocol}
The primary unit of analysis is a completed run identified by agent,
model, worldline, and seed. Every reported series---agent NAV, holdings,
the equal-weight baseline, and every figure in this paper---shares one
horizon: each run terminates at 2035-12-31, with AC NAV reconstructed
\emph{daily} from each executed rebalance's post-trade account value and
target weights marked on the worldline panel closes and anchored at the
horizon's terminal account value, and FM and baseline series marked
daily. Every run is assigned a status of \texttt{complete},
\texttt{partial}, \texttt{running}, \texttt{pending}, or
\texttt{failed}; partial and running artifacts are excluded from
headline comparisons. Released tables carry a row-level manifest so that
missing combinations are visible rather than silently dropped.

For drawdown, $\mathrm{DD}_t=V_t/\max_{s\leq t}V_s-1$ and
$\mathrm{MDD}=\min_t\mathrm{DD}_t$. Cumulative transaction cost is
audited per run as $C_{\mathrm{bps}}=10^4\sum_t c\,\Delta_t/V_0$ over
one-way migrated notional $\Delta_t$ at rate $c{=}3$\,bps, exactly the
cost leg of the gate contract of \S\ref{sec:agents-adapt}. Robustness is
summarized across the nine worldlines with the median, interquartile
range, minimum, and maximum of each metric; statistical intervals and
paired tests are computed from completed run units only and never treat
decision blocks from one worldline as independent replications. A
configuration field is marked \texttt{observed} only when present in
the recorded request or runner---never substituted from an official
provider default---so comparisons are not silently based on
undocumented defaults.

\paragraph{Execution and fill conventions.}
At a decision date $t$, every observation is masked to dates $\le t$,
i.e., through the close of $t$. FM emits its target after the observation
finalizes and fills at the next trading mark; since the forward panels
carry no overnight gap (each forward open equals the previous close,
Appendix~\ref{app:worldlines}), FM's fill price is the last close it
observed. AC's native trader fills at the open of the decision day
itself---the close of $t-1$---so the change in weights it executes is
exposed to the $t{-}1\to t$ close-to-close span, whose endpoint the agent
had already observed; positions are then marked at the close of $t$. We
quantify this same-day overlap for every AC run by applying each executed
rebalance's weight change to its decision day's close-to-close return and
summing over the horizon: the cumulative effect averages $-0.8$
percentage points of initial capital (per-run range $[-4.8,+4.6]$,
positive in only 5 of 18 runs)---sign-inconsistent and small relative to
the reported return spread, so no reported ranking or comparison changes.
The deterministic references decide at the grid close and fill at the next
open under the same one-way migrated-notional cost contract; EW-15 therefore
uses no signal input beyond its fixed equal target.

%% file: sections/appendix_worldlines.tex
\clearpage
\section{Worldline Panels, Generator Details, and Data Quality}
\label{app:worldlines}

This appendix records the artifact-level details behind the benchmark of
\S\ref{sec:worldlines}: panel construction, the full stage table, and
the input-quality boundaries released with the panels.

\paragraph{Panel construction.}
Each of the nine final panels carries 83{,}347 daily rows over
2020-01-01--2035-12-31 with no duplicated (asset, date) keys; the forward
policy information boundary is 2026-07-15; the observed common execution/market
anchor is 2026-07-16; and synthetic divergence begins on 2026-07-17 and
continues through 2035-12-31 (2{,}468 daily marks per WL). Within each worldline, the corresponding market panel is byte-identical across
the four configurations. All nine online panels regenerate from the released realized-history panel and
English machine-readable scenario specification to floating-point tolerance.
WL9's undated third narrative stage introduces no numeric endpoint; its next
dated anchor remains explicit in the specification.

\paragraph{Input-quality boundaries.}
Released alongside the panels, rather than silently patched, are two
input-quality boundaries: seven online-synthesized series (CN10Y, DXY,
EURUSD, SOX, US10Y, USDCNY, USDJPY) carry empty high/low fields where the
warmup window ends in zero-amplitude bars, and weekend stage endpoints
(e.g., WL7's 2029-06-30) are anchors, not tradable dates. Open/close
consumers are unaffected; full-OHLC consumers must treat those fields as
absent.

\paragraph{Forward observation schema.}
Forward (post-boundary) rows are fully determined by the generated close
path and by pre-boundary statistics; Table~\ref{tab:schema} states the
rule per field. The forward panel therefore contains no information
channel beyond the close path: open, high, low, volume, amount, and vwap
are deterministic functions of the closes and of warmup statistics, so
factor expressions built on them degenerate to close-derived signals in
the forward period (vwap equals the close exactly). The yield series are
synthetic quoted-yield exposures. Their returns are defined as percentage
changes in the quoted yield level; they are not bond total-return or
duration-scaled series. Their level paths preserve the worldline's basis-point
moves anchored at the boundary level, volume is zero, and high/low are
undefined as noted above.

\begin{table}[H]
\centering
\small
\begin{tabular}{@{}lll@{}}
\toprule
Field & Warmup (real data) & Forward (synthetic) \\
\midrule
close & observed & generator output (Eq.~\ref{eq:gbb}, re-anchored) \\
open & observed & previous close \\
high / low & observed & $\max/\min(\text{open},\text{close})\times(1\pm\mathrm{rng}/2)$ \\
volume & observed & per-asset warmup median (constant) \\
amount & observed & close $\times$ volume \\
vwap & amount $/$ volume & equals close \\
PE, PS, PB, DYR & observed & undefined forward \\
\bottomrule
\end{tabular}
\caption{\textbf{Observation schema, warmup vs.\ forward.} rng is the
per-asset median of $(\text{high}-\text{low})/\text{close}$ over the last
60 pre-boundary days; high/low are undefined for the seven zero-amplitude
series listed above, and volume is zero (hence vwap undefined) for the
yield, FX, and VIX series.}
\label{tab:schema}
\end{table}

\paragraph{Scenario provenance artifacts.}
The released English JSON attaches nine stage narratives as provenance metadata
to the numeric dates and multi-asset endpoints consumed by the package-local
generator. It also freezes asset order, parameters, and a versioned CRC32 seed
contract: asset seeds use \texttt{worldline\_number|asset\_id}, then segment
seeds use \texttt{asset\_seed|segment\_index}. Three unified 2026-07 baseline anchors (000300.SH,
SPX, HSI) are designated Suntime platform values and the remainder are
market-checked estimates. The agent-team debate is documented as scenario
origination rather than as an independently executable platform export; its
post-debate score matrix is reported in Appendix~\ref{app:provenance}.

\paragraph{Full stage table.}
Table~\ref{tab:stages} lists, for each worldline, the stage terminal
dates and narratives parsed from the machine-readable waypoint file that
accompanies each panel; Table~\ref{tab:families} in the main text is its
compact form.

\begin{table}[H]
\centering
\scriptsize
\setlength{\tabcolsep}{2.6pt}
\begin{tabular}{@{}ll@{\ }p{5.9cm}@{}}
\toprule
WL & Stage end dates & Stage narratives (in order) \\
\midrule
WL1 & 27-12, 28-05, 29-12, 31-12, 35-12 & geopolitical escalation: red-line warnings $\to$ 72-hour blitz $\to$ sanctions \& global restructuring $\to$ new equilibrium $\to$ new normal \\
WL2 & 27-12, 29-12, 31-12, 35-12 & internet splintering $\to$ fintech-system splintering $\to$ bipolar consolidation $\to$ bipolar digital economy \\
WL3 & 27-12, 28-06, 29-12, 31-12, 35-12 & nuclear test escalation $\to$ conventional conflict $\to$ deterrence failure \& restructuring $\to$ non-proliferation rebuild $\to$ new nuclear order \\
WL4 & 27-12, 29-12, 31-12, 35-12 & cracks spreading $\to$ Minsky moment $\to$ stagflation \& de-dollarization $\to$ early institutional rebuild \\
WL5 & 27-12, 29-12, 31-12, 35-12 & cracks emerge $\to$ JPY disorderly-devaluation spiral $\to$ Asian currency fragmentation $\to$ Japan's ``new normal'' \\
WL6 & 27-12, 29-12, 31-12, 35-12 & inflation re-ignition $\to$ central-bank credibility collapse $\to$ entrenched stagflation $\to$ new monetary order \\
WL7 & 27-12, 29-06, 30-12, 32-12, 35-12 & fragility build-up \& first resonance $\to$ homogenization culminating in an algorithmic run on banks $\to$ regulatory rebuild $\to$ AI-regulation paradigm $\to$ new equilibrium \\
WL8 & 27-12, 29-12, 31-12, 33-12, 35-12 & Middle-East escalation $\to$ Hormuz blockade \& energy-system collapse $\to$ violent energy restructuring $\to$ new-energy order $\to$ end of the fossil era \\
WL9 & 27-12, 29-12, ---, 33-12, 35-12 & pre-pandemic growth $\to$ XPV-27 outbreak \& labor collapse $\to$ deep recession, automation sprouts$^\dagger$ $\to$ automation/AI supercycle $\to$ digital-economy normal \\
\bottomrule
\end{tabular}
\caption{\textbf{The nine worldlines: full stage table.} Stage terminal
dates (YY-MM) and narratives, represented in the machine-readable JSON
specification. WL9's third stage carries no explicit numeric date
($^\dagger$; its narrative lies between
2029-12 and 2033-12). All WLs end 2035-12-31.}
\label{tab:stages}
\end{table}

%% file: sections/appendix_significance.tex
\clearpage
\section{Window-Level Persistence Analysis}
\label{app:significance}

Section~\ref{sec:results-ew} compares agents against the EW-15
baseline on terminal returns, which yields only nine paired comparisons per
agent column. This appendix repeats the comparison at decision granularity as
a \emph{temporal persistence diagnostic}: it tests whether the baseline's
advantage arises only from a few terminal episodes or accumulates steadily
across the horizon. The shared ten-trading-day decision grid partitions each
run into $245$ non-overlapping windows; for each window we compare the agent's
realized window return (net of costs, from the daily NAV series of
Appendix~\ref{app:metrics}) with EW-15's window return over the same dates, a
\emph{win} being a strictly higher agent return. This yields $2{,}205$ paired
comparisons per agent column and $8{,}820$ in total. Windows within a worldline
are serially dependent---strategy state and market regimes persist across
windows---so the intervals below use a descriptive cluster bootstrap that
resamples the nine fixed worldlines with replacement ($10^{4}$ resamples,
fixed seed). Because these worldlines are a fixed scenario set rather than
draws from a population, the intervals are descriptive scenario-resampling
summaries rather than population-level significance claims.

These window-level persistence diagnostics are secondary to the
retrospective-to-prospective rank-transfer analysis in
Section~\ref{sec:results-rr}; they are reported descriptively rather than as
the main empirical narrative.

\begin{table}[h]
\centering
\small
\setlength{\tabcolsep}{5pt}
\begin{tabular}{@{}lcc@{}}
\toprule
Agent column & Window hit rate & Mean excess / window \\
 & (\%, descriptive interval) & (bps, descriptive interval) \\
\midrule
AC-terra & 45.0 $[43.5,\,46.6]$ & $-8.2$ $[-9.0,\,-7.2]$ \\
AC-DS & 47.4 $[45.4,\,49.5]$ & $-5.7$ $[-9.9,\,-2.1]$ \\
FM-terra & 49.9 $[49.0,\,50.8]$ & $+2.3$ $[-2.0,\,+6.9]$ \\
FM-DS & 48.7 $[46.5,\,51.0]$ & $-10.3$ $[-15.0,\,-6.0]$ \\
\midrule
Pooled & 47.8 $[46.7,\,48.9]$ & $-5.5$ $[-7.0,\,-4.0]$ \\
\bottomrule
\end{tabular}
\caption{\textbf{Per-window comparison against EW-15} (245
non-overlapping ten-trading-day windows per run; 8,820 paired comparisons).
Hit rate is the share of windows the agent beats EW-15; excess is the agent's
window return minus EW-15's, in basis points. Intervals are descriptive
worldline-clustered bootstrap summaries.}
\label{tab:window}
\end{table}

Three columns have hit rates below one half and negative mean excess; their
descriptive intervals exclude zero (Table~\ref{tab:window}). FM--terra is the
exception: its hit rate is near one half and its positive point estimate for
mean window excess ($+2.3$ bps) has an interval spanning zero, consistent
with its $4/9$ endpoint record. AC-terra's negative mean window excess is
directionally consistent with its endpoint shortfall. The window-level view
also shows when the baseline wins, while remaining descriptive for the fixed
scenario set.

%% file: sections/appendix_rr.tex
\clearpage
\section{Native-Policy Retrospective Replay}
\label{app:rr}

\paragraph{Definition and interpretation.}
Native-Policy Retrospective Replay (RR) freezes each configuration's learned
boundary policy at \texttt{2026-07-15} and replays it over the realized history from
which it was obtained. In compact form,
\[
D_{2020:2026}\rightarrow S_c^{\mathrm{learned}}(2026)
\rightarrow \operatorname{Replay}(S_c^{\mathrm{learned}},D_{2020:2026}).
\]
``Native'' is scoped to the learned policy and proposal-generation mechanism:
each framework-specific signal and target-weight policy is executed from
archived native artifacts. Retrospective proposals are then evaluated under a
common causal close$(t)\!\to$open$(t+1)$ execution contract. The policy state
includes frozen policy code, factor/library artifacts, window and mapper
configuration, and source version. RR is deliberately retrospective and
in-sample: the final learned policy is replayed on the same realized history
from which it was obtained, so its returns are not estimates of future
performance. The test asks whether the ranking induced by realized-history
evaluation agrees with rankings under unseen prospective worldlines. It does
not identify historical familiarity, in-sample selection, or training-data
contamination as separate causal mechanisms.

\paragraph{Frozen state and replay contract.}
The four configurations are AC--terra, AC--DS, FM--terra, and FM--DS. FM--DS
uses the validated canonical eight-factor boundary state. Exact boundary target reconstruction
is verified against the archived first-decision references of all nine
worldlines. No LLM is called and no factor is re-mined or re-selected during
replay. The realized panel begins on \texttt{2020-01-06}; an independent
readiness pass gives the common \texttt{t\_start=2020-07-20}, followed by 157
ten-trading-day decisions. Every policy starts from zero holdings and
1,000,000 initial capital. Signals observe through $\operatorname{close}(t)$,
the official deterministic gate evaluates the proposal, and fills occur at
$\operatorname{open}(t+1)$. Portfolios are long-only, fully invested, and use
the fixed 15-asset order. One-way migrated notional costs 3 bps; the first
allocation is free and a rejected proposal leaves holdings unchanged.

\paragraph{Exact boundary verification.}
All four frozen boundary policies pass an archived first-decision golden
comparison: target asset IDs and order are exact, weights sum to one, and
$\texttt{max\_abs\_diff}=0$, $\texttt{L1}=0$, and Spearman $\rho=1$.
This is an exact boundary check, not a claim that a historical trajectory
reproduces an unavailable original learning run. The complete replay records
finite-weight, long-only, asset-order, next-open, gate, cost, and fresh-runtime
invariants. Two complete runs produce
byte-identical CSV, JSON, Markdown, and PNG outputs (\texttt{audit/hashes.json}).
Source paths and SHA256 values are provided in the anonymous artifact package.

\begin{table}[H]
\centering
\scriptsize
\setlength{\tabcolsep}{3pt}
\begin{tabular}{l l r r r}
\toprule
Configuration & Boundary reference & max abs & L1 & $\rho$ \\
\midrule
AC--terra & archived native first decision & 0 & 0 & 1 \\
AC--DS & archived native first decision & 0 & 0 & 1 \\
FM--terra & archived native first decision & 0 & 0 & 1 \\
FM--DS & nine archived WL references & 0 & 0 & 1 \\
\bottomrule
\end{tabular}
\caption{Native boundary reconstruction golden tests. Exact comparisons use
the fixed asset set and order; source and bundle hashes are recorded in the
artifact manifest.}
\label{tab:rr-golden}
\end{table}

\begin{table}[H]
\centering
\scriptsize
\setlength{\tabcolsep}{2.5pt}
\begin{tabular}{lrrrrrrrrrr}
\toprule
Configuration & Cum. & Ann. & Vol. & Sharpe & MDD & Proposals & Executed & Proposal turnover & Cost (bps of $V_0$) & Agent rank \\
\midrule
FM--DS & 275.88\% & 21.64\% & 17.25\% & 1.223 & -23.56\% & 157 & 156 & 68.40 & 499.0 & 1 \\
AC--terra & 220.51\% & 18.81\% & 12.01\% & 1.496 & -12.43\% & 157 & 8 & 22.42 & 1.8 & 2 \\
FM--terra & 192.15\% & 17.19\% & 15.45\% & 1.105 & -32.04\% & 157 & 156 & 71.23 & 389.0 & 3 \\
AC--DS & 174.03\% & 16.09\% & 10.52\% & 1.471 & -11.89\% & 157 & 145 & 15.65 & 77.4 & 4 \\
EW--15 & 227.47\% & 19.19\% & 14.41\% & 1.292 & -22.38\% & 157 & 157 & 3.60 & 19.0 & --- \\
\bottomrule
\end{tabular}
\caption{Full RR metrics on the common realized-history window. Cum. is
terminal return, Ann. is annualized return, Vol. is annualized volatility,
and MDD is maximum drawdown. Proposal turnover sums one-way proposed
migration, including gated-out proposals; executed counts and costs refer
only to accepted rebalances. Cost is in basis points of initial benchmark
capital. RR and prospective cumulative returns have different horizons and
are not compared by raw magnitude.}
\label{tab:rr-full}
\end{table}

\begin{table}[H]
\centering
\scriptsize
\begin{tabular}{r r r r}
\toprule
WL & Spearman $\rho$ & Preserved pairs & FM--DS WL rank \\
\midrule
1 & -0.8 & 1/6 & 3 \\
2 & -0.8 & 1/6 & 4 \\
3 & -0.8 & 1/6 & 4 \\
4 & -0.4 & 2/6 & 2 \\
5 & -0.8 & 1/6 & 4 \\
6 & +0.8 & 5/6 & 1 \\
7 & +0.4 & 4/6 & 2 \\
8 & -1.0 & 0/6 & 4 \\
9 & -0.4 & 2/6 & 4 \\
\midrule
\multicolumn{4}{l}{Mean $\bar\rho=-0.42$; median $\rho=-0.80$; retention $17/54=31.5\%$;}\\
\multicolumn{4}{l}{RR leader = FM--DS; top-1 survival $1/9$; FM--DS median rank 4.}\\
\bottomrule
\end{tabular}
\caption{Prospective rank transfer for the four agent configurations. Rankings
are defined by terminal cumulative return within each environment. Statistics
are descriptive: no p-values or significance tests are applied.}
\label{tab:rr-rank}
\end{table}

\paragraph{Metric sensitivity.}
Return ranking is primary because terminal return is the benchmark's headline
outcome. A Sharpe-based sensitivity analysis yields mean $\rho\approx0.00$,
median $-0.20$, and 50\% pairwise retention; hence the precise magnitude of
rank mismatch is metric-dependent, while stable retrospective-to-prospective
ordering is not recovered.

\begin{table}[H]
\centering
\scriptsize
\begin{tabular}{lrrrr}
\toprule
Configuration & RR excess sign & WLs $>$ EW & Median WL excess & Median WL rank \\
\midrule
AC--terra & negative & 0/9 & -2.14 pp & 3 \\
AC--DS & negative & 1/9 & -0.94 pp & 2 \\
FM--terra & negative & 4/9 & -0.41 pp & 1 \\
FM--DS & positive (+2.46 pp) & 0/9 & -2.41 pp & 4 \\
\bottomrule
\end{tabular}
\caption{Apparent excess return relative to each environment's EW-15
reference. The RR and prospective environments use different horizons, so
only within-environment excess and signs are interpreted.}
\label{tab:rr-ew}
\end{table}

\paragraph{Reproducibility artifacts.}
The accompanying anonymous artifact package contains \texttt{config.json},
\texttt{frozen\_state\_manifest.json}, the golden, provenance, validation,
and hash manifests, complete daily NAV series and decision records, and the
scripts used to verify and plot the derived results. The deterministic
acceptance command is \texttt{rr\_replay.py --twice}; its recorded result is
\texttt{byte\_identical=true}.

%% file: sections/appendix_factors.tex
\clearpage
\section{Staged Factor Inventory}
\label{app:factors}

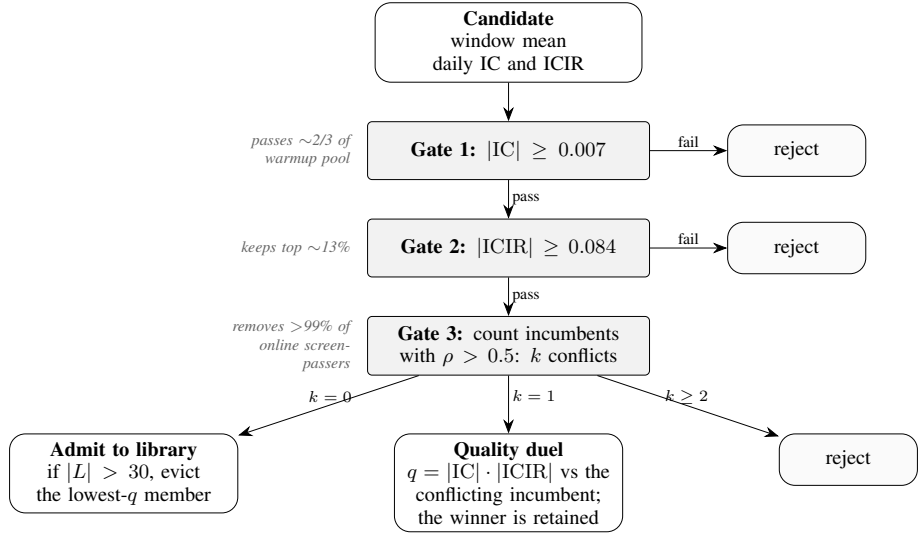
\begin{figure}[H]
\centering
\resizebox{0.86\columnwidth}{!}{%
\begin{tikzpicture}[
  font=\footnotesize,
  node distance=6mm and 12mm,
  gate/.style={draw, rounded corners=2pt, align=center, fill=gray!10,
               inner sep=4pt, text width=4.1cm, minimum height=9mm},
  term/.style={draw, rounded corners=7pt, align=center,
               inner sep=3.5pt, text width=3.9cm, minimum height=8mm},
  rej/.style={term, fill=gray!4, text width=1.9cm},
  arr/.style={-{Stealth[length=2.2mm]}, thin},
  lab/.style={font=\scriptsize, inner sep=1.5pt},
  rate/.style={font=\scriptsize\itshape, black!60, inner sep=1.5pt,
               align=right, text width=2.6cm}
]
\node[term] (cand) {\textbf{Candidate}\\ window mean daily IC and ICIR};
\node[gate, below=of cand] (g1) {\textbf{Gate 1:} $|\mathrm{IC}|\ge 0.007$};
\node[gate, below=of g1] (g2) {\textbf{Gate 2:} $|\mathrm{ICIR}|\ge 0.084$};
\node[gate, below=of g2] (g3) {\textbf{Gate 3:} count incumbents\\ with $\rho>0.5$: $k$ conflicts};
\node[rej, right=of g1] (r1) {reject};
\node[rej, right=of g2] (r2) {reject};
\node[rej, below right=9mm and 42mm of g3.south] (r3) {reject};
\node[term, below=9mm of g3, text width=3.3cm] (duel) {\textbf{Quality duel}\\ $q=|\mathrm{IC}|\cdot|\mathrm{ICIR}|$ vs the
  conflicting incumbent; the winner is retained};
\node[term, below left=9mm and 42mm of g3.south, text width=3.3cm] (adm) {\textbf{Admit to library}\\ if $|L|>30$, evict the
  lowest-$q$ member};
\node[rate, left=2mm of g1] (a1) {passes $\sim$2/3 of\\ warmup pool};
\node[rate, left=2mm of g2] (a2) {keeps top $\sim$13\%};
\node[rate, left=2mm of g3] (a3) {removes $>$99\% of\\ online screen-\\passers};
\draw[arr] (cand) -- (g1);
\draw[arr] (g1) -- node[lab, right] {pass} (g2);
\draw[arr] (g1) -- node[lab, above] {fail} (r1);
\draw[arr] (g2) -- node[lab, right] {pass} (g3);
\draw[arr] (g2) -- node[lab, above] {fail} (r2);
\draw[arr] (g3) -- node[lab, right, pos=0.35] {$k\ge 2$} (r3);
\draw[arr] (g3) -- node[lab, right, pos=0.35] {$k=1$} (duel);
\draw[arr] (g3) -- node[lab, left, pos=0.35] {$k=0$} (adm);
\end{tikzpicture}}
\caption{\textbf{The admission funnel.} Gate values are stipulated and
deliberately permissive; selectivity comes from the staged tightening.
Gate 3 counts the candidate's pairwise Spearman conflicts
($\rho>0.5$) with incumbents: $k\ge2$ rejects the candidate; $k=1$
triggers a quality duel on $q=|\mathrm{IC}|\cdot|\mathrm{ICIR}|$ with
the conflicting incumbent, the winner retained; $k=0$ admits it, with
the capacity-30 trim evicting the lowest-$q$ member. Gray rates are
measured on the warmup candidate pool and on online screen-passers.}
\label{fig:admission}
\end{figure}

The factor lists are staged by lifecycle phase: warmup libraries are shared
across the nine worldlines of an experiment and would otherwise be repeated
ninefold, so we separate (i) each experiment's warmup qualified library
(2020-01-01--2026-07-15), (ii) factors admitted online into a terminal
live library (at most capacity 30), and (iii) fast-screen-passed mining
candidates that never passed the in-library correlation gate. Table~\ref{tab:lifecycle}
summarizes the counts; Tables~\ref{tab:factors-examples}--\ref{tab:factors-online}
give representative factors with their exact formulas; the complete
machine-readable inventory is released as described below.

\begin{table}[h]
\centering
\footnotesize
\setlength{\tabcolsep}{4pt}
\begin{tabular}{@{}>{\raggedright\arraybackslash}p{3.1cm}>{\raggedright\arraybackslash}p{2.1cm}>{\raggedright\arraybackslash}p{2.1cm}>{\raggedright\arraybackslash}p{2.0cm}>{\raggedright\arraybackslash}p{2.1cm}@{}}
\toprule
 & AC--terra & AC--DS & FM--terra & FM--DS \\
\midrule
Warmup qualified library (enters every WL) & 6 & 4 & 21 & 8 \\
Online-admitted terminal members listed per WL$^{*}$ & 270 (30 per WL; all distinct) & 99 (3--20 per WL; 94 distinct) & 11 (on 7 WLs) & 0 \\
Terminal live-library size across WLs & 30 (at capacity) & 0--22 & 21--24 & 8 (warmup only) \\
Mining-screen candidates ($\Sigma \ge 3$), never admitted & --- & --- & --- & 109 ($\Sigma$ 824) \\
\bottomrule
\end{tabular}
\caption{\textbf{Factor lifecycle in numbers.} $^{*}$AC--DS's WL8 is the
citation-listed exception: its library was evicted to zero before the
horizon (\S\ref{sec:results-process}). FM--DS admitted nothing online;
its 109 unadmitted candidates accumulated $\Sigma=824$ screenings across
the nine mining streams. The reported inventory summaries regenerate from the released
factor-lifecycle and factor-library audit logs.}
\label{tab:lifecycle}
\end{table}

\begin{table}[h]
\centering
\footnotesize
\setlength{\tabcolsep}{4pt}
\begin{tabular}{@{}>{\raggedright\arraybackslash}p{1.7cm}>{\raggedright\arraybackslash}p{2.8cm}>{\raggedright\arraybackslash}p{1.2cm}>{\raggedright\arraybackslash}p{7.0cm}@{}}
\toprule
Experiment & Factor & IC / ICIR & Formula \\ \midrule
AC--terra & \texttt{peer\_\allowbreak{}median\_\allowbreak{}leadlag\_\allowbreak{}5d} & +0.056 / 0.17 & \texttt{median(\allowbreak{}\{return\_\allowbreak{}5d[j] for j != i\})}\\
AC--terra & \texttt{miner\_\allowbreak{}3\_\allowbreak{}clv\_\allowbreak{}1d} & +0.068 / 0.19 & \texttt{-(\allowbreak{}2 * (\allowbreak{}close - low) /\allowbreak{} (\allowbreak{}high - low) - 1)}\\
AC--DS & \texttt{mom\_\allowbreak{}10d\_\allowbreak{}skip5} & +0.041 / 0.12 & \texttt{close.shift(\allowbreak{}5) /\allowbreak{} close.shift(\allowbreak{}15) - 1.0}\\
AC--DS & \texttt{vix\_\allowbreak{}beta\_\allowbreak{}cond\_\allowbreak{}60x20} & -0.038 / -0.09 & \texttt{-beta(\allowbreak{}asset\_\allowbreak{}ret,\allowbreak{} VIX\_\allowbreak{}ret,\allowbreak{} 60) * (\allowbreak{}VIX/\allowbreak{}VIX.shift(\allowbreak{}20) - 1.0)}\\
FM--terra & \texttt{return\_\allowbreak{}rank\_\allowbreak{}mean\_\allowbreak{}reversion} & +0.031 / 0.09 & \texttt{Neg(\allowbreak{}CsRank(\allowbreak{}TsRank(\allowbreak{}Sum(\allowbreak{}\$returns,\allowbreak{} 5),\allowbreak{} 30)))}\\
FM--terra & \texttt{intraday\_\allowbreak{}vwap\_\allowbreak{}pressure} & +0.028 / 0.09 & \texttt{CsZScore(\allowbreak{}Mul(\allowbreak{}Div(\allowbreak{}Sub(\allowbreak{}\$close,\allowbreak{} \$vwap),\allowbreak{} Add(\allowbreak{}Abs(\allowbreak{}\$vwap),\allowbreak{} 0.001)),\allowbreak{} Div(\allowbreak{}Sub(\allowbreak{}\$vwap,\allowbreak{} \$open),\allowbreak{} Add(\allowbreak{}Abs(\allowbreak{}\$open),\allowbreak{} 0.001))))}\\
FM--DS & \texttt{decayed\_\allowbreak{}return\_\allowbreak{}signal} & +0.033 / 0.09 & \texttt{CsRank(\allowbreak{}Decay(\allowbreak{}\$returns,\allowbreak{} 10))}\\
FM--DS & \texttt{cumulative\_\allowbreak{}amount\_\allowbreak{}per\_\allowbreak{}share} & +0.028 / 0.11 & \texttt{CsZScore(\allowbreak{}Div(\allowbreak{}CumSum(\allowbreak{}\$amt),\allowbreak{} Add(\allowbreak{}CumSum(\allowbreak{}\$volume),\allowbreak{} 1)))}\\
\bottomrule
\end{tabular}
\caption{\textbf{Representative warmup factors} (IC/ICIR from the warmup
evaluation; each library seeds all nine worldlines of its experiment).
The complete lists of 6/4/21/8 factors are in the released manifest.}
\label{tab:factors-examples}
\end{table}

\begin{table}[h]
\centering
\footnotesize
\setlength{\tabcolsep}{4pt}
\begin{tabular}{@{}>{\raggedright\arraybackslash}p{2.2cm}>{\raggedright\arraybackslash}p{2.9cm}>{\raggedright\arraybackslash}p{2.2cm}>{\raggedright\arraybackslash}p{5.2cm}@{}}
\toprule
Stage / stream & Factor & Evidence & Formula \\ \midrule
AC--terra, WL1 & \texttt{miner\_\allowbreak{}1\_\allowbreak{}20281116\_\allowbreak{}defensive\_\allowbreak{}relative\_\allowbreak{}lead\_\allowbreak{}20d} & cited 29-04-19$\sim$35-12-21 (144 dec.) & \texttt{(\allowbreak{}\$asset\_\allowbreak{}return\_\allowbreak{}20d - mean(\allowbreak{}XAU,\allowbreak{} US10Y,\allowbreak{} CN10Y 20d)) * (\allowbreak{}0.25 + 0.75*I(\allowbreak{}breadth\_\allowbreak{}20d < 0.40)),\allowbreak{} lagged}\\
AC--terra, WL1 & \texttt{macro\_\allowbreak{}stress\_\allowbreak{}resilience\_\allowbreak{}20d} & cited 29-12-13$\sim$35-12-21 (100 dec.) & \texttt{shift1(\allowbreak{}(\allowbreak{}r20 - median\_\allowbreak{}cs(\allowbreak{}r20)) /\allowbreak{} (\allowbreak{}downside\_\allowbreak{}vol20 + 0.02) * (\allowbreak{}1 + 0.35*clip(\allowbreak{}(\allowbreak{}VIX-median60(\allowbreak{}VIX))/\allowbreak{}std60(\allowbreak{}VIX),-1,2)*tanh(\allowbreak{}5*(\allowbreak{}r20-median\_\allowbreak{}cs(\allowbreak{}r20)))))}\\
AC--DS, WL2 & \texttt{spx\_\allowbreak{}corr60} & cited 26-08-13$\sim$35-12-06 (198 dec.) & \texttt{close.pct\_\allowbreak{}change(\allowbreak{}).rolling(\allowbreak{}60).corr(\allowbreak{}close['SPX'].pct\_\allowbreak{}change(\allowbreak{}))}\\
AC--DS, WL3 & \texttt{cn10y\_\allowbreak{}beta\_\allowbreak{}60} & cited 26-07-30$\sim$35-12-20 (227 dec.) & \texttt{rolling\_\allowbreak{}beta(\allowbreak{}asset\_\allowbreak{}daily\_\allowbreak{}ret,\allowbreak{} cn10y\_\allowbreak{}yield\_\allowbreak{}daily\_\allowbreak{}change,\allowbreak{} 60)}\\
FM--terra, WL4 & \texttt{range\_\allowbreak{}adjusted\_\allowbreak{}return\_\allowbreak{}rank} & admitted 2026-08-01 & \texttt{CsRank(\allowbreak{}Div(\allowbreak{}TsRank(\allowbreak{}\$returns,\allowbreak{} 13),\allowbreak{} Add(\allowbreak{}TsRank(\allowbreak{}Div(\allowbreak{}Sub(\allowbreak{}\$high,\allowbreak{} \$low),\allowbreak{} Add(\allowbreak{}Abs(\allowbreak{}\$close),\allowbreak{} 0.001)),\allowbreak{} 13),\allowbreak{} 0.001)))}\\
FM--terra, WL5 & \texttt{volume\_\allowbreak{}acceleration\_\allowbreak{}pressure} & admitted 2026-08-03 & \texttt{CsZScore(\allowbreak{}Delta(\allowbreak{}EMA(\allowbreak{}Log(\allowbreak{}Add(\allowbreak{}\$volume,\allowbreak{} 1)),\allowbreak{} 6),\allowbreak{} 14))}\\
FM--DS (mining) & \texttt{return\_\allowbreak{}residual\_\allowbreak{}on\_\allowbreak{}volume} & $\Sigma = 43$ & \texttt{CsRank(\allowbreak{}Resid(\allowbreak{}\$returns,\allowbreak{} \$volume,\allowbreak{} 20))}\\
FM--DS (mining) & \texttt{dema\_\allowbreak{}ema\_\allowbreak{}momentum} & $\Sigma = 33$ & \texttt{CsRank(\allowbreak{}Sub(\allowbreak{}DEMA(\allowbreak{}\$close,\allowbreak{} 10),\allowbreak{} EMA(\allowbreak{}\$close,\allowbreak{} 20)))}\\
FM--DS (mining) & \texttt{kama\_\allowbreak{}cross\_\allowbreak{}signal} & $\Sigma = 17$ & \texttt{CsRank(\allowbreak{}Sub(\allowbreak{}KAMA(\allowbreak{}\$close,\allowbreak{} 20),\allowbreak{} KAMA(\allowbreak{}\$close,\allowbreak{} 5)))}\\
\bottomrule
\end{tabular}
\caption{\textbf{Representative online admissions and mining candidates.}
AC spans show first--last decision citing the factor ($n$ decisions);
FM--terra gives the admitting WL and timestamp; mining rows give the total
screen count $\Sigma$ across the nine FM--DS mining streams (all
failed the correlation gate; none was ever admitted).}
\label{tab:factors-online}
\end{table}

\subsection{Complete machine-readable inventory}

The tables above are an excerpt. The released manifest carries every
factor admitted by any run---id, exact formula, category, warmup IC/ICIR,
per-decision citation spans, and admission/eviction timestamps with
machine-readable reasons---in the per-run factor-lifecycle logs and
library audit trails shipped with the trajectories. One gap is exposed,
not hidden: ten AC--terra ids cited by decisions have no retained
artifact, and the manifest marks them as missing. The full tables (all
6/4/21/8 warmup factors, the 270, 99, and 11 online memberships per WL,
and all 109 supplement candidates with formulas) and every count of
Table~\ref{tab:lifecycle} regenerate from the released package-local factor
libraries, lifecycle logs, and decision traces using
\texttt{scripts/reproduce\_factor\_inventory.py}. The script writes the
summary CSV, lifecycle CSV, and compact appendix table under
\texttt{reproduced/factor\_inventory/}.

%% file: sections/appendix_process.tex
\clearpage
\section{Process Registers and the WL8 Forensic Note}
\label{app:process}

\S\ref{sec:results-process} evaluates each run's coherence across three
states---research state, declared decision state, executed state. This
appendix records the observables and registers behind that evaluation
and reconstructs the WL8 degeneration at mechanism level. It is a
forensic record rather than a new result: every fact regenerates from
the released decision, ensemble, and execution traces.

\subsection{Process observables and registers}
\label{app:process-obs}

Each run is summarized by five trajectory-level observables:
\begin{equation}
\begin{aligned}
L_t &= |\mathcal L_t|,\\
C_t &= \text{admissions}_t+\text{evictions}_t,\\
A_t &= \text{admissions}_t,\\
D_t &= \|w_t^{\mathrm{declared}}-w_t^{\mathrm{executed}}\|_1,\\
F_t &= \|w_t-w_t^{\mathrm{fallback}}\|_1.
\end{aligned}
\label{eq:observables}
\end{equation}
where $L_t$ is the live factor-library size, $C_t$ is the count of admissions
plus evictions in the decision window, and $A_t$ is the admission count,
$D_t$ the distance between the portfolio weights the agent \emph{records}
and those it \emph{executes}, and $F_t$ the distance of executed weights
from the equal-weight fallback policy. These observables are derived from the
released factor-lifecycle, ensemble, decision, and holdings traces. Table~\ref{tab:process} complements them with the
terminal values of the raw registers from which they derive: the two
event ledgers (\emph{dec} and \emph{reb}, below), cumulative transaction
cost, and terminal library size versus ensemble selection.

\begin{table}[H]
\centering
\small
\setlength{\tabcolsep}{3.4pt}
\begin{tabular}{@{}lrrrr@{}rrrr@{}}
\toprule
 & \multicolumn{4}{c}{AC-terra} & \multicolumn{4}{c}{AC-DS} \\
\cmidrule(lr){2-5}\cmidrule(lr){6-9}
WL & dec & reb & cost & lib$\to$ens & dec & reb & cost & lib$\to$ens \\
\midrule
1 &  43 & 205 & 30.9 & 30$\to$8  &  35 & 212 & 100.3 & 14$\to$6 \\
2 & 121 & 159 & 33.8 & 30$\to$9  &  47 & 200 & 102.8 & 22$\to$5 \\
3 & 294 & 134 & 31.1 & 30$\to$7  &  19 & 228 &  71.8 & 22$\to$10 \\
4 &  98 &  45 & 16.1 & 30$\to$5  &  64 & 187 &  45.2 & 3$\to$2 \\
5 &  76 & 159 & 37.8 & 30$\to$7  & 235 &  12 &   3.7 & 12$\to$0 \\
6 & 157 &  90 & 43.1 & 30$\to$6  &  42 & 244 & 125.8 & 15$\to$8 \\
7 & 174 &  30 &  4.2 & 30$\to$6  & 255 &  16 &   3.8 & 8$\to$7 \\
8 & 185 &  57 & 18.0 & 30$\to$5  & 202 &  47 &   7.9 & \textbf{0}$\to$3 \\
9 & 120 & 121 & 26.8 & 30$\to$10 & 100 & 135 &  30.2 & 20$\to$10 \\
\bottomrule
\end{tabular}
\caption{\textbf{AC process registers.} dec = declined proposals:
each cycle's proposed portfolio shift is ruled on by comparing its
expected gross edge against friction (migration) cost, and every dec
entry is a rejection by that ruling (tagged
\texttt{gross\_edge\_not\_above\_migration\_cost}, unchanged target, or
missing forecast); each entry carries the proposed weights, forecasts,
and factor ids. reb = executed
rebalances: the initial build plus every proposal the gate accepted,
each with its transferred notional and realized cost. The two registers
partition the proposal stream under the shared gate (see below); cost = cumulative transaction
cost in bps of initial capital over 9.5 years (3\,bps on migrated
notional, 0 on the initial build); lib$\to$ens = terminal live-library
size $\to$ ensemble-selected factors. FM terminal libraries hold 21--24
factors (terra) and exactly the 8 warmup factors (DS) in every WL; the
staged inventory---warmup libraries, online admissions, and the DS
mining supplement---is in Appendix~\ref{app:factors}.}
\label{tab:process}
\end{table}

\paragraph{One gate, two ledgers.} The migration gate of
\S\ref{sec:agents-adapt} is the single ruling instance in both
frameworks, and the two registers record its two outcomes. A rejected
proposal is appended to \emph{dec} with its skip reason and holdings are
untouched: all 43 proposals in AC-terra WL1, all 174 in WL7, and all 294
in WL3 were declined, which is why $\mathrm{dec}$ can exceed
$\mathrm{reb}$ without any trade having occurred. A proposal the gate
accepts executes and is appended to \emph{reb} with its transferred
notional and realized cost. The only executions outside the gate are the
initial build (once per run) and the executor's self-healing
full-investment repair, which is tagged as such and never occurred in
the released runs; $\mathrm{reb}\gg\mathrm{dec}$ therefore marks cycles
that proposed and were approved, never trades that bypassed a ruling.
Recording both outcomes separately keeps such divergences visible; the
WL8 forensic below dissects the extreme case of a run that mostly
declined to trade.

\subsection{Forensic note: anatomy of the silent degeneration}
\label{app:process-forensic}

\S\ref{sec:results-process} reports the AC--DS WL8 degeneration in
outline; this note reconstructs the mechanism-level chain that the row
$0\to 3$ in Table~\ref{tab:process} summarizes.

\paragraph{Research state: the library evicted to zero ($L_t\to 0$).}
The live factor library was progressively emptied by the admission
contract's own quality arbitration: each eviction removed the
lower-$q$ member of a conflicting pair, so the library contracted as the
agent shed factors that no longer cleared the diversity/quality bar. In
the agent's own decision logic this is deliberate pruning---actively
deleting factors whose quality no longer meets the bar, to cut losses
from weak signals early rather than let them drag the portfolio; a
majority of audit cycles already recorded kept${}=0$ well before the
horizon, so the terminal empty library was not an end-state accident
but the endpoint of a sustained trajectory. The pruning nonetheless
left the agent without live signals.

\paragraph{Declared state: a frozen, unreachable ensemble ($D_t>0$).}
Decision records nonetheless kept citing a three-factor ensemble. The
cited file matched a hardcoded fallback ensemble last written in 2031
and was never updated; a second ensemble, written to a different
directory, never reached the trader. The writer/reader path divergence
means the declared portfolio---built from the three factor
ids---corresponded to no executed configuration:
$w_t^{\mathrm{declared}}\neq w_t^{\mathrm{executed}}$ on the remaining
horizon.

\paragraph{Executed state: exact equal weight ($F_t\to 0$).}
With the preference mapper's 22\% single-asset cap binding on every one
of the fifteen assets, all weights clip and renormalize to exactly
$1/15$. For nearly the entire horizon the run therefore held a basket
numerically identical to the EW-15 policy while its records showed three
factor ids---execution had behaviorally converged to the fallback and
$F_t=0$ on those intervals. On WL8 that policy returned $+93.2\%$
against the run's $+86.3\%$.

\paragraph{Reading the row.}
Table~\ref{tab:process} lists the WL8 AC--DS row as $0\to 3$ (terminal
live-library size $\to$ ensemble-selected factors) across 202 recorded
decision entries and 47 executed rebalances. The three registers this
note dissects---research state, declared state, and realized
holdings---are precisely the ones the evaluation contract records
separately, which is what makes the dissociation visible at all.

%% file: sections/appendix_provenance.tex
\clearpage
\section{Scenario Provenance: Post-Debate Review Scores}
\label{app:provenance}

The scenario set of \S\ref{sec:worldlines} was produced by the debate
workflow of \S\ref{sec:worldlines-audit}. Each scenario was
independently derived by one of the team's three members---strategy
(policy) / macro-finance / chief analyst (synthesis)---and then
cross-reviewed by the full team through three debate rounds, scoring
every scenario on logical rigor, an agent-elicited 10-year plausibility range, and
shock severity. Table~\ref{tab:reviews-full} reports the full score
matrix. The scores document how the scenario set was produced; the
benchmark itself assigns no likelihood weights.

\begin{table}[h]
\centering
\small
\setlength{\tabcolsep}{5pt}
\begin{tabular}{@{}lcccc@{}}
\toprule
WL & Theme & Rigor (/5) & 10-yr plaus.\ (\%) & Severity (/10) \\
\midrule
1 & geopolitical escalation & 3 & 10--15 & 9 \\
2 & tech ``silicon curtain'' & 4 & 40--50 & 7.5 \\
3 & nuclear-proliferation chain & 3 & 8--12 & 8.5 \\
4 & Treasury ``Minsky moment'' & 3.5 & 18--25 & 10 \\
5 & JPY devaluation spiral & 3 & 15--22 & 7 \\
6 & stagflation \& CB collapse & 3 & 10--15 & 8 \\
7 & algorithmic run on banks & 3.5 & 18--25 & 8 \\
8 & energy-system collapse & 4 & 12--18 & 9 \\
9 & pandemic supply-chain rebuild & 4 & 10--18 & 9.5 \\
\bottomrule
\end{tabular}
\caption{\textbf{Post-debate review scores per worldline}, documenting the
scenario workflow: logical rigor (5-point
scale), agent-elicited 10-year plausibility range, and shock severity
(10-point scale).}
\label{tab:reviews-full}
\end{table}

%% file: sections/appendix_models.tex
\clearpage
\section{Model Configuration}
\label{app:models}

All 36 runs share the configuration of Table~\ref{tab:models}. Both models use
standard reasoning mode; gpt-5.6-terra uses medium reasoning effort and
DeepSeek uses high reasoning effort. Temperature and $top_p$ remain at provider
defaults for both.

\begin{table}[H]
\centering
\small
\begin{tabular}{@{}lllll@{}}
\toprule
Model & Reasoning mode & Reasoning effort & Temperature & Top-p \\
\midrule
gpt-5.6-terra & standard & medium & default & default \\
deepseek-v4-flash & standard & high & default & default \\
\bottomrule
\end{tabular}
\caption{\textbf{LLM configuration for all runs.}}
\label{tab:models}
\end{table}